\documentclass[fleqn,10pt]{wlscirep}
\usepackage[utf8]{inputenc}
\usepackage[T1]{fontenc}
\usepackage{cite}
\usepackage{amsmath,amssymb,amsfonts}
\usepackage{algorithmic}
\usepackage{graphicx}
\usepackage{subfigure}
\usepackage{caption}
\usepackage{textcomp}
\usepackage{hyperref}
\usepackage{multicol}
\usepackage{multirow}
\usepackage{interval}
\usepackage{adjustbox}

\title{QAmplifyNet: Pushing the Boundaries of Supply Chain Backorder Prediction Using Interpretable Hybrid Quantum--Classical Neural Network}

\author[1]{Md Abrar Jahin}
\author[2]{Md Sakib Hossain Shovon}
\author[1]{Md. Saiful Islam}
\author[3,*]{Jungpil Shin}
\author[2]{M. F. Mridha}
\author[3]{Yuichi Okuyama}

\affil[1]{Department of Industrial Engineering and Management, Khulna University of Engineering and Technology (KUET), Khulna 9203, Bangladesh}
\affil[2]{Department of Computer Science, American International University-Bangladesh (AIUB), Dhaka 1229, Bangladesh}
\affil[3]{Department of Computer Science and Engineering, The University of Aizu, Aizuwakamatsu 965-8580, Japan}

\affil[*]{Corresponding author: Jungpil Shin (e-mail: jpshin@u-aizu.ac.jp).}

\keywords{Hello, Keyword2, Keyword3}

\begin{abstract}
Supply chain management relies on accurate backorder prediction for optimizing inventory control, reducing costs, and enhancing customer satisfaction. Traditional machine-learning models struggle with large-scale datasets and complex relationships. This research introduces a novel methodological framework for supply chain backorder prediction, addressing the challenge of collecting large real-world datasets with 90\% accuracy. Our proposed model demonstrates remarkable accuracy in predicting backorders on short and imbalanced datasets. We capture intricate patterns and dependencies by leveraging quantum-inspired techniques within the quantum-classical neural network QAmplifyNet. Experimental evaluations on a benchmark dataset establish QAmplifyNet's superiority over eight classical models, three classically stacked quantum ensembles, five quantum neural networks, and a deep reinforcement learning model. Its ability to handle short, imbalanced datasets makes it ideal for supply chain management. We evaluate seven preprocessing techniques, selecting the best one based on Logistic Regression's performance on each preprocessed dataset. The model's interpretability is enhanced using Explainable Artificial Intelligence techniques. Practical implications include improved inventory control, reduced backorders, and enhanced operational efficiency. QAmplifyNet also achieved the highest F1-score of 94\% for predicting "Not Backorder" and 75\% for predicting "Backorder," outperforming all other models. It also exhibited the highest AUC-ROC score of 79.85\%, further validating its superior predictive capabilities. QAmplifyNet seamlessly integrates into real-world supply chain management systems, empowering proactive decision-making and efficient resource allocation. Future work involves exploring additional quantum-inspired techniques, expanding the dataset, and investigating other supply chain applications. This research unlocks the potential of quantum computing in supply chain optimization and paves the way for further exploration of quantum-inspired machine learning models in supply chain management. Our framework and QAmplifyNet model offer a breakthrough approach to supply chain backorder prediction, offering superior performance and opening new avenues for leveraging quantum-inspired techniques in supply chain management. 
\end{abstract}

\begin{document}

\flushbottom
\maketitle
%
%
\thispagestyle{empty}

\section*{Introduction}

\label{sec:introduction}

Supply chain management (SCM) plays a critical role in ensuring the smooth flow of goods and services from manufacturers to end consumers. In this context, accurate prediction of backorders, which refers to unfulfilled customer orders due to temporary stockouts, is of paramount importance. Supply chain backorder prediction (SCBP) enables proactive inventory management, efficient resource allocation, and enhanced customer satisfaction. It assists in mitigating the negative impacts of stockouts, such as lost sales, decreased customer loyalty, and disrupted production schedules. Predicting backorders for products in the future is challenging, mainly because the demand for a particular product can fluctuate unexpectedly. To develop an accurate predictive model, it is crucial to have an adequate amount of training data derived from the inventory tracking system. This data allows the model to learn the patterns that indicate whether a product will likely be backordered. However, a significant challenge in building such a model is the inherent imbalance in the dataset. The number of samples where a product is backordered is much lower than those where products are not backordered. This class imbalance creates a skewed dataset, which can negatively impact the model's performance.

There needs to be more research available on product backordering, specifically addressing the challenges of class imbalance \cite{santis2017predicting, hajek2020profit}. However, extensive work has been conducted in the past to optimize inventory management. Inventory managers encounter various challenges when faced with material shortages, which can result in complete backlogs or lost orders. Previous literature has categorized material backordering as fixed, partial, or time-weighted backorders \cite{srivastav2016multi}. Customers' willingness to wait for a replenished stock depends on factors such as supplier reputation, recency of the backorder placement, and waiting time. Some customers may be patient and wait, while others may seek alternative options due to impatience. In such cases, the supplier experiences sales loss and missed revenue opportunities, leading to customer dissatisfaction and potential doubts about the supplier's inventory management capabilities.

Traditional prediction models, predominantly based on classical machine learning (CML) algorithms, have been widely utilized for backorder prediction. However, these models face several challenges when dealing with large-scale datasets typically encountered in supply chain (SC) applications. Traditional models often need help handling these datasets' complexity and dimensionality, leading to suboptimal performance and limited scalability. Moreover, the ability to capture intricate patterns and dependencies within the data is crucial for accurate prediction, which remains a challenge for conventional approaches. Despite the widespread use of CML models, tuning millions of hyperparameters during training CML models like DNNs requires significant computing power. The fast-rising data volume required for training, particularly in the post-Moore’s Law era, exceeds the limit of semiconductor production technology, which limits the field's advancement. On the other hand, quantum computing (QC) has proven to be more effective at solving issues that are insurmountable for conventional computers, such as factoring big numbers and doing unstructured database searches \cite{harrow}. Nevertheless, because of the noise produced by the quantum gates and the absence of quantum error correction on Noisy Intermediate Scale Quantum (NISQ) devices, QC with substantial circuit depth faces significant difficulties. So, creating quantum algorithms with a reasonable level of noise-resistant circuit depth would be of fundamental relevance. The performance of CML models is now being outperformed by quantum machine learning (QML), which is based on variational quantum circuits \cite{biamonte}. The vastly decreased number of model parameters is one of the key advantages of variational quantum models over their classical counterparts. As a result, variational quantum models reduce the overfitting issues related to CML. Moreover, under some circumstances, they may learn more quickly or attain better test accuracy compared to their conventional counterparts \cite{chen}. The variational quantum model plays a vital role as the quantum component of a modern QML architecture, with the circuit parameters being updated by a classical computer \cite{mitarai}.

The emergence of quantum-inspired techniques has opened up new avenues for addressing the limitations of CMLs \cite{zidan2023}. These techniques, inspired by QC principles, leverage the inherent parallelism and quantum-inspired optimization algorithms to enhance predictive capabilities. QML models exhibit promising potential in handling large-scale datasets, capturing complex patterns, and improving prediction accuracy in various domains. SCBP can benefit from enhanced model performance, improved accuracy, and more efficient resource allocation by harnessing the power of quantum-inspired techniques. The utilization of QML algorithms can enable the identification of intricate relationships between variables, facilitating more accurate prediction of backorder occurrences. Consequently, these techniques have the potential to optimize SCM, minimize stockouts, reduce costs, and enhance customer satisfaction.

In the field of inventory management research, numerous studies have delved into enhancing forecasting and decision-making regarding backorders. A multitude of these studies have explored various mathematical and algorithmic approaches, such as Markov decision processes, reinforcement learning (RL), fuzzy models, and ML techniques. However, a conspicuous research gap exists when it comes to effectively handling short and imbalanced datasets, a common real-world scenario where acquiring extensive data is often impractical. While ensemble forecasting models have demonstrated superiority over other methods, their computational efficiency becomes a limiting factor, particularly with large warehouse datasets, thus restricting their practical applicability. In light of the limitations of traditional ML models and the potential advantages offered by quantum-inspired techniques, this research aims to develop a novel Hybrid Quantum-Classical Keras Neural Network (NN) tailored for SCBP to bridge this research gap. The proposed model combines the flexibility and interpretability of Keras NNs with quantum-inspired optimization algorithms to overcome the limitations of classical approaches. By integrating quantum-inspired techniques into the prediction process, we anticipate achieving improved accuracy, robustness, and scalability in SCBP.

The novelty of this research lies in the application and benchmarking of QML techniques against the classical ML techniques in the field of SCBP for short and imbalanced datasets. Extensive literature review and analysis confirm that this study represents the first known instance of QML implementation in the context of SCBP. By introducing QML to this domain, our objective is to unlock new possibilities and harness the potential advantages that quantum-inspired techniques can offer to the field of SCM. This research contributes to the field of SCM by exploring the potential of QML techniques for accurate and efficient backorder prediction. A novel hybrid Quantum-Classical Neural Network (Q-CNN) was developed as part of this study, combining the strengths of parallel-processed NN computing and quantum physics. Hybrid classical-quantum computing is a computational paradigm that combines classical infrastructure with quantum computers to address specific problem domains. In this approach, classical computers play a crucial role in pre-processing data, controlling quantum machines during computation, and post-processing the results obtained from quantum computations. By harnessing quantum phenomena such as entanglement and superposition, quantum computers possess the ability to perform parallel processing in a manner unprecedented by classical computers \cite{zidan2021}. By leveraging the strengths of both classical and quantum computing, hybrid systems enable the utilization of quantum resources while utilizing classical algorithms and techniques to enhance overall computational performance. This synergistic combination allows for the efficient utilization of quantum resources and the effective integration of classical and quantum computing capabilities to tackle complex problems. The hybrid algorithms employed in this study outperformed their classical counterparts by leveraging quantum and classical computing capabilities.

In light of these considerations, this research provides a novel and thorough methodology for anticipating inventory backorders. The goal is to maximize profits while minimizing costs related to product backorders, maintaining good relationships with suppliers and customers, and preventing sales from being lost. Customers and businesses alike can profit from precise projections of future backorders for individual products with the help of a well-developed predictive model. A current topic of research is the simplification of quantum algorithms for usage with NISQ computers \cite{preskill2018quantum}. Quantum algorithms that scale well may be efficiently executed on computers that use photons, superconductors, or trapped ions \cite{schuld2019quantum, havlicek2019supervised,johri2021nearest}. Particularly exciting is QML because of its compatibility with current NISQ designs \cite{hubregtsen2022training, saggio2021experimental}. Predicting product backorders, for example, requires access to massive amounts of data, which is a strength of traditional ML algorithms. For this reason, this research introduces a novel Q-CNN model that can deal with data imbalances even when trained on a small dataset. NISQ devices are effective in running shallow-depth algorithms requiring a few qubits \cite{preskill2018quantum}. Given the specific difficulties and prerequisites of product backordering prediction, it becomes sensitive to take advantage of QML run on NISQ devices by means of the SCBP dataset. Classification in inspection tests for small-size datasets was made possible by searching for a quantum advantage on the classifier \cite{tomono2022performance}.

\begin{figure*}[!ht]
  \centering

  \subfigure[Before Removal of Null Values]{
    \includegraphics[width=0.8\textwidth,height=0.35\textheight]{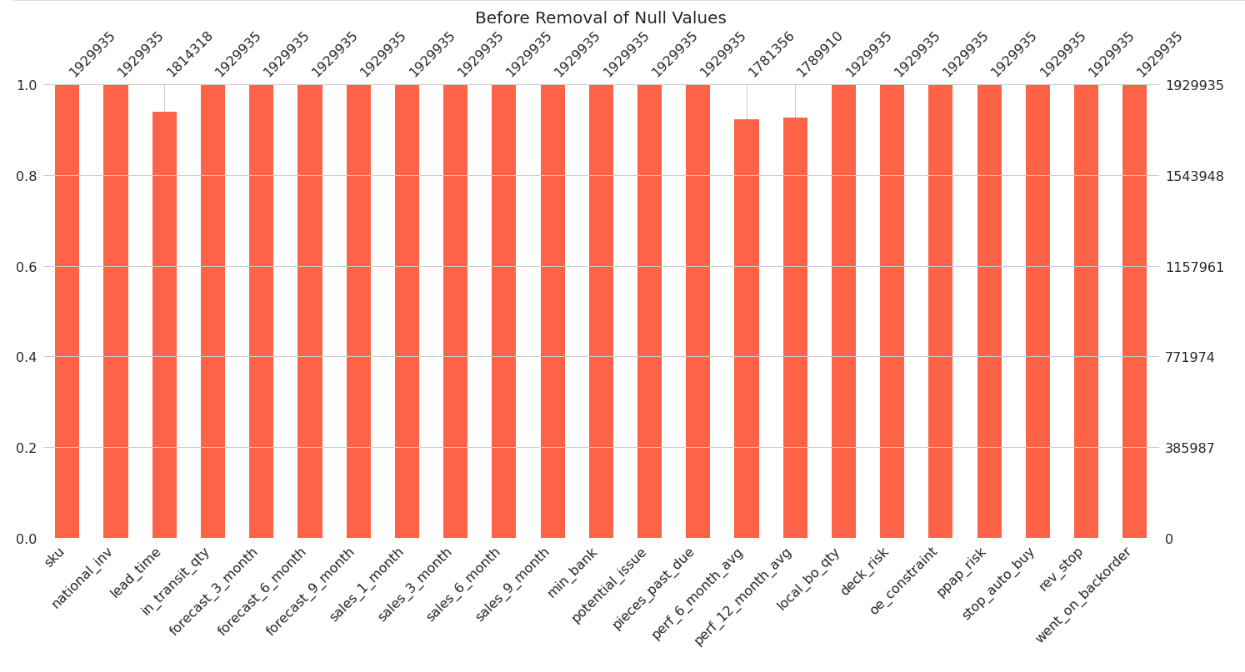}
    \label{fig1:subfig1}
  }
  \hfill
  \subfigure[After Removal of Null Values]{
    \includegraphics[width=0.8\textwidth,height=0.35\textheight]{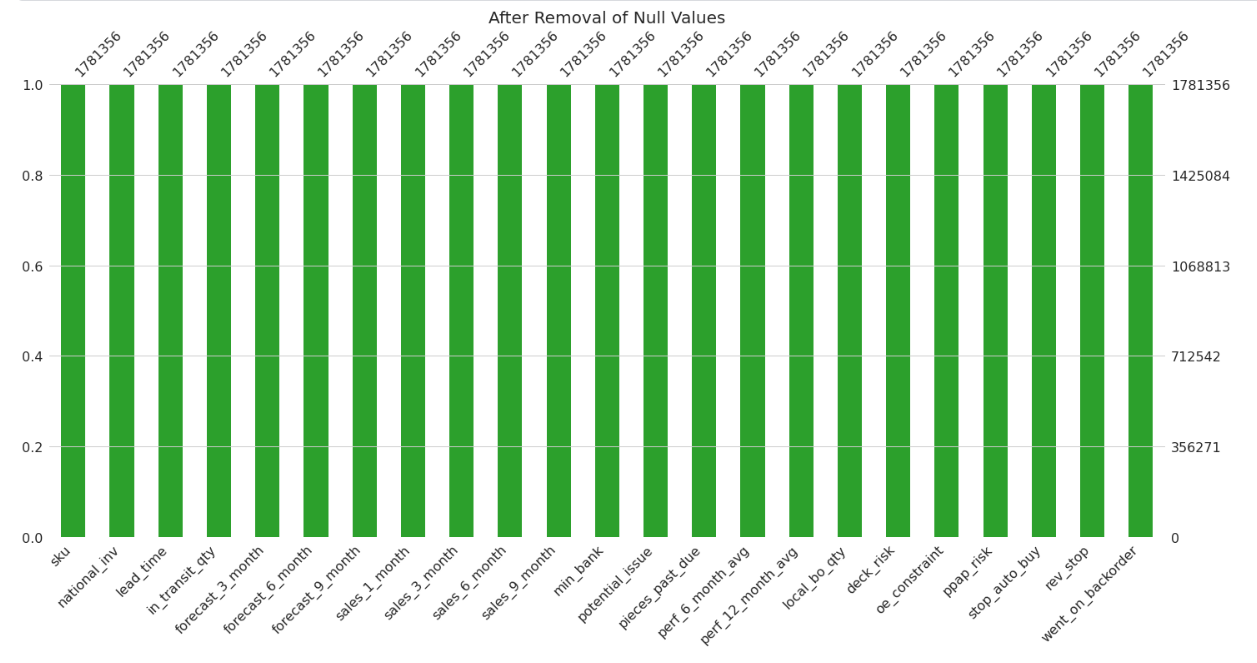}
    \label{fig1:subfig2}
  }

  \caption{Barplots showing the distribution of null values in the dataset (a) before and (b) after removal. The top of each bar shows the number of samples present in each feature.}
  \label{fig1}
\end{figure*}

The open-access Kaggle dataset used in this research was gathered from an 8-week inventory management system \cite{santis2017predicting}. Unfortunately, as shown in Figure \ref{fig1}, the dataset needs to be balanced because the number of backordered items is disproportionately high (137:1). Figure \ref{fig2} shows a dataset heatmap, indicating that high feature correlations are required, increasing the difficulty of working with the dataset. The issue is made more complicated by the fact that any prediction model will need help dealing with imbalanced datasets \cite{li2017backorder}. Gradient boosting model (GBM), random forest (RF), and logistic regression (LR) are only some of the traditional ML methods that have been presented for similar jobs in the past \cite{santis2017predicting, hajek2020profit}. It has also been common practice to use undersampling and oversampling strategies to rectify grossly unbalanced business statistics \cite{kang2014robust,garcia2018dynamic}.

This research presents an innovative approach for SCBP that incorporates effective preprocessing methods, resulting in a novel quantum-classical ML-based prediction model. There are various steps in the methodological flowchart. We first preprocess each SKU's features using seven possible combinations of methods. We benchmark each preprocessed dataset by applying the LR model and then select the most effective preprocessing technique based on its accuracy. The selected preprocessing tasks involve converting categorical features into numerical features, handling missing values, log transforming numerical features, normalizing feature values within a specific range, and dropping redundant numerical features using variation inflation factor (VIF) treatment. In this classification problem, there are substantially fewer positive samples (backordered) than negative samples (non-backordered). Consequently, we address the issue of class imbalance by employing an undersampling technique called NearMiss. We choose undersampling instead of oversampling because QML models struggle to train on large datasets compared to CML models. Furthermore, we utilize principal component analysis (PCA) to extract four input principal components from the preprocessed dataset. These components capture the most significant features for prediction.

Finally, we propose our hybrid Q-CNN model, named QAmplifyNet, which incorporates key aspects of the architecture in its mnemonic name. The "Q" signifies the utilization of QC principles, highlighting the model's quantum component. "Amplify" represents the concept of amplifying information through the model's layers. Lastly, "Net" refers to the NN nature of the model, incorporating both classical and QML components. For the performance evaluation of our model, we compare it against eight commonly used CML models, one deep RL model, five quantum NNs, and three quantum-classical stacked ensembles. Through this comprehensive comparison, we aim to demonstrate the superiority and robustness of our proposed QAmplifyNet model for SCBP on short datasets. Despite the excellent accuracy of CML models on this complete dataset, our proposed QAmplifyNet model holds significant value. It showcases remarkable performance on short, imbalanced data, which is a common challenge in SC inventory management. Additionally, the application of QML in this domain represents a pioneering effort, making it the first-ever QML application in the SC inventory management field.

Using a benchmark SCBP dataset titled \textit{"Can you predict product backorder?"}, we run tests utilizing the proposed model. The experimental findings demonstrate the higher performance of our technique in SCBP, as evaluated by accuracy and area under the receiver operating characteristic (ROC) curve. Moreover, we compare our models to well-known classification models and come to the conclusion that our strategy performs noticeably better than other comparable models.

\begin{figure*}[!ht]
  \centering
  \includegraphics[width=1\textwidth, height=0.9\textwidth]{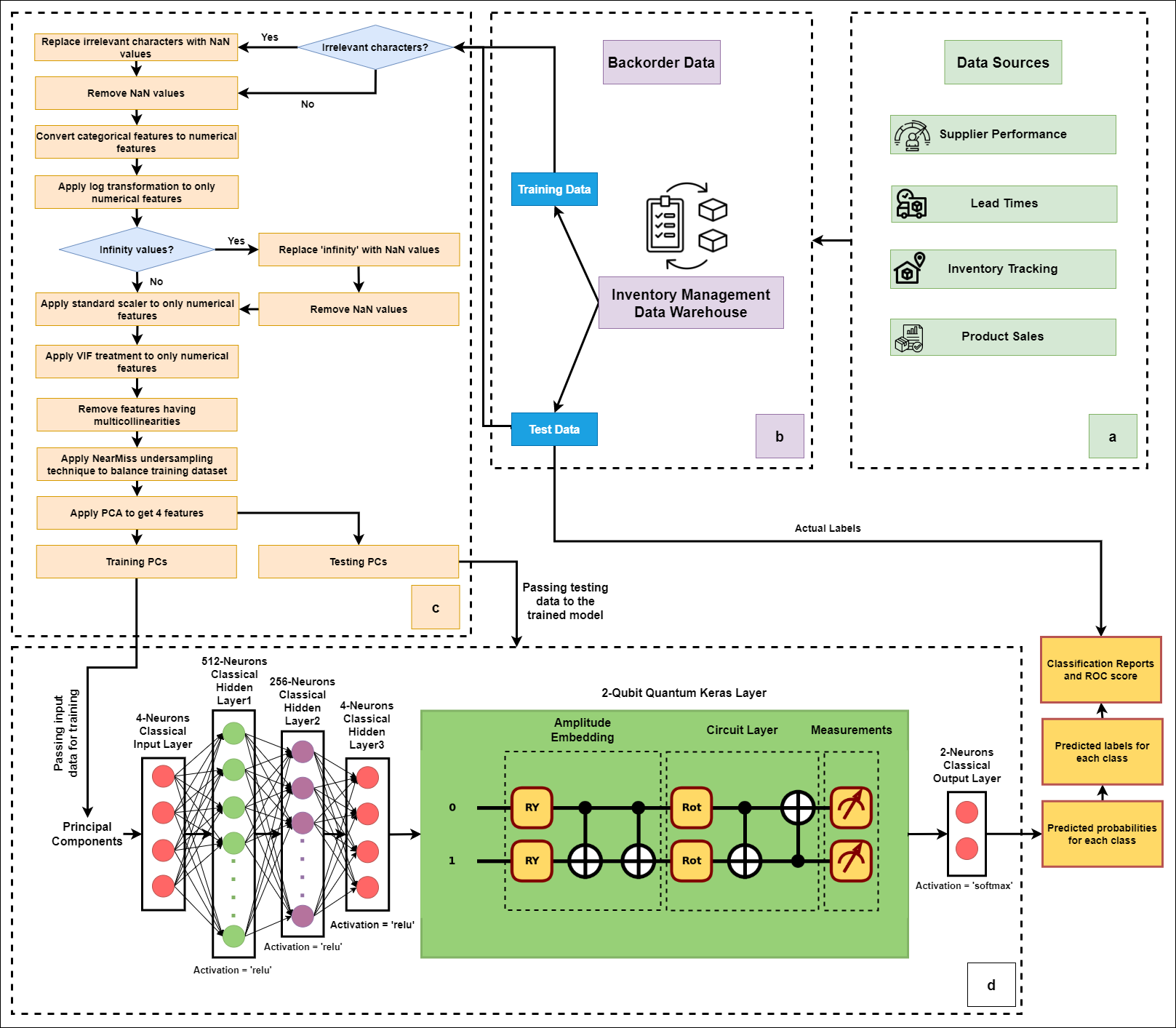}
  \caption{Methodological framework illustrating (a) data sources, (b) data collection and splitting, (c) data preprocessing, and (d) proposed Q-CNN model development for SCBP.}
  \label{proposed_method}
\end{figure*}

In summary, this paper makes eight-fold contributions:

\begin{enumerate}

    \item This study represents the pioneering application of QC in the SCM domain.
    \item We introduce a novel theoretical framework for predicting inventory backorders.
    \item We present a comprehensive data preprocessing technique that combines log transformation, normalization, VIF treatment, and NearMiss undersampling to address the imbalanced nature of the dataset in the rare SCM domain.
    \item We propose a hybrid quantum-classical Keras NN-based technique for forecasting product backorders, enhancing the suppliers' overall efficiency.
    \item We demonstrate that the hybrid Q-CNN model overcomes the challenge of limited availability of large SCM datasets by showcasing its enhanced performance compared to CML and QML models on short datasets with few features.
    \item We enhance the interpretability of the proposed model by implementing Explainable Artificial Intelligence (XAI) methods, specifically SHAP and LIME.
    \item Our novel methodology significantly improves prediction accuracy, reducing misclassification errors, especially false positives and false negatives, and ultimately increasing enterprise profitability. 
    \item Lastly, we discuss how the proposed methodology can be applied homogeneously to other supervised binary classification domains, such as predicting credit card defaults.
\end{enumerate}

Here's how the paper breaks down: the current literature on SCBP, CML models, quantum-inspired models, and RL-based techniques are reviewed in Section ``\hyperref[sec:rw]{Related Work}". It draws attention to the unanswered questions that our proposed model seeks to answer. Section ``\hyperref[sec:methods]{Methods}" introduces the proposed hybrid Q-CNN-based backorder prediction system to address class imbalance on the short dataset. It describes the selected preprocessing steps followed by the architectures and working principles of the models used in this paper. In Section ``\hyperref[sec:results]{Results}", we use the experimental data and robustness tests to evaluate, compare, and verify the effectiveness of the proposed model. In Section ``\hyperref[sec:discussion]{Discussion}", we conclude and make comparisons between the proposed model and alternative methods. Possible real-world SCM implementations of the suggested method are also discussed. The report finishes with a summary of the main findings and the main contributions in Section ``\hyperref[sec:conclusion]{Conclusion and Future Work}". Some future research directions in SCBP employing quantum-inspired approaches are also presented.

\section*{Related Work}
\label{sec:rw}
In the field of scientific research on inventory management, various studies have been conducted to improve forecasting and decision-making related to backorders. \cite{mart2013tactical} proposed a solution based on Markov decision processes to define inventory and backorder strategies. They treated production system yield as a stochastic process. \cite{xu2017finite} examined a stock inventory system incorporating periodic reviews and a partial backorder forecast. They developed a framework considering the distribution of demand and its factors to assess uncertainty in the inventory system. \cite{prak2019general} analyzed estimating errors and derived an inventory model's predicted lead time demand distribution. This distribution could be used to optimize inventory management. \cite{chaharsooghi2008reinforcement} determined ordering policies in inventory systems using RL. They viewed SCM as a multi-agent system and utilized the Q-learning technique to solve the model. \cite{abdul2009two} combined the N-retailer problem and overall cost considerations to develop an objective function for ordering, storing, and backordering in a single inventory. They optimized three decisions jointly: lot sizing, routing and distribution, and replenishment.

\cite{brahimi2016multi} developed an integrated multi-item manufacturing and routing problem model considering constrained warehousing and manufacturing capabilities. For optimal lot-sizing, routing and distribution, and replenishment, they factored in a penalty cost for backordered products. \cite{van2018base} highlighted the significance of accurately calculating backorder levels to enhance fill rates and reduce total costs. \cite{ghiami2016planning} considered partial backordering and discounts on backordered products in a production-inventory system to maximize net present value. \cite{ganesh2019multi} emphasized the importance of incorporating backorder decisions and costs into an ideal inventory policy, noting that previous models often overlooked SCBP. \cite{bjork2009analytical} introduced fuzzy number-based optimization models to account for uncertain demand and lead times, outperforming conventional methods. \cite{kazemi2015incorporating} presented a fuzzy model that included human reasoning with backorders, while \cite{lin2010economic} and \cite{taleizadeh2012economic} constructed economic ordering quantity models with various factors such as special sale pricing, poor quality, partial backordering, and quantity discounts. \cite{kim2018improved} proposed an integrated inventory model that optimized multiple decisions simultaneously, including lead time, lot size, number of shipments, and safety factor. In a fuzzy condition, \cite{kazemi2010inventory} analyzed a warehouse model incorporating backorders using fuzzy numbers and a graded mean integration model. \cite{guo2019decisions} optimized spare component allocation decisions for serviceable standalone systems with dependent backorders. \cite{feng2012demand} devised an approach to forecasting order for line-replaceable unit components with backorders, highlighting the need to consider the dynamic features of these factors. \cite{shin2012development} proposed a framework to reduce overall costs and anticipated risk costs of backorder replenishment plans using a Bayesian belief network. \cite{wang2014dynamic} investigated a dynamic rationing scheme that considered demand dynamics, while \cite{bao2018decomposition} used a Markov decision support system to determine the best rationing levels across all categories of demand. \cite{trapero2019empirical,trapero2019quantile} developed non-parametric and parametric prediction models, such as kernel density and GARCH algorithms, to predict safety stock and reduce long lead times. \cite{santis2017predicting} proposed ensemble-based machine learning algorithms, GBM and RF paired with undersampling, for SCBP. \cite{hajek2020profit} discussed the benefits and limitations of ensemble prediction methods and undersampling in dealing with noisy data and improving prediction accuracy. \cite{liu2022} improved SCBP with the use of the Conditional Wasserstein Generative Adversarial Network (CWGAN) model along with Randomized Undersampling (RUS). Initially, the majority of the non-backorder samples were reduced using RUS. Second, CWGAN was used as a technique for oversampling to provide superior backorder samples. Ultimately, RF was implemented to predict backorders. The class imbalance problem was successfully addressed by \cite{shajalal} densely linked DNNs, which combined SMOTE and randomized undersampling. The experimental outcomes indicate better prediction performance and predicted profit on a thorough product backordering dataset, proving the proposed model's superiority over existing ML approaches.

In handling noisy data and minimizing overfitting, ensemble forecasting models have shown superiority to non-parametric and parametric forecasting methods. However, their computational efficiency becomes a limitation when analyzing large warehouse datasets in real time, limiting their practical utility. On the other hand, undersampling techniques can enhance computational performance, but they may also exclude potentially valuable training data and compromise prediction accuracy. To address these challenges, we propose a hybrid Q-CNN applied to a short backorder dataset; our preprocessing approach involves several steps. Firstly, we apply a log transform to the data, followed by standard scaling to normalize the features. We also address multicollinearity issues by implementing Variable Inflation Factor (VIF) treatment. With the training dataset being unbalanced, we employ the NearMiss undersampling method, which involves deliberately reducing the majority of class occurrences.

The choice of a hybrid Q-CNN for analyzing the short dataset is driven by its unique advantages, primarily rooted in the challenging nature of SCBP with limited data. In real-world SCM, acquiring extensive datasets is a formidable task due to the hesitance of companies to disclose their internal data. Collecting vast data is often infeasible, considering the significant time and effort required. Given these constraints, our approach focuses on developing QAmplifyNet, a model specifically designed to predict backorders accurately, even when trained on relatively short datasets. We balanced and shortened the preprocessed dataset using the NearMiss algorithm after selecting the optimal preprocessing technique to accomplish this. Despite the inherent challenges associated with short datasets, our hybrid Q-CNN demonstrates remarkable predictive accuracy. By offering these insights, we aim to elucidate the unique advantages of our approach when dealing with the inherent limitations of short SCBP datasets. Combining classical and QC techniques, this approach harnesses the power of quantum algorithms for specific tasks while leveraging the robustness and versatility of CML frameworks like Keras. Exploiting quantum principles like superposition and entanglement through the use of quantum-inspired algorithms inside a classical NN framework can result in efficient and more accurate calculations \cite{zidan2018}. Compared to purely classical or purely quantum models, the hybrid Q-CNN is anticipated to outperform in several aspects. Firstly, the combination of classical and quantum techniques enables more powerful computations, leading to increased accuracy in backorder predictions. The utilization of quantum-inspired algorithms within the classical framework allows for more efficient exploration of the solution space and better identification of patterns and trends in the data. The hybrid approach offers practical advantages over pure quantum models. Quantum computers are still in the early stages of development, and their availability and scalability may pose limitations in real-world applications. CML models often grapple with scalability issues when handling extensive datasets, resulting in computational bottlenecks that hinder their practicality. The intricate relationships and dynamic nature of SC data constrain their real-time prediction capabilities. These limitations become particularly evident when striving for timely SCBPs. Our research highlights the imperative need for advanced solutions by quantum-inspired techniques, such as our QAmplifyNet model, to transcend these constraints, ultimately enhancing the accuracy and efficiency of real-time predictions while addressing the scalability challenges inherent to pure classical or pure quantum models. 

The hybrid model can leverage existing computational resources and infrastructures by integrating CML frameworks like Keras, making it more accessible and practical for implementation in real-world SC environments. This integration allows for more accurate predictions, improved decision-making, and better inventory control, making it a promising approach for addressing the challenges of backorder management in real-world contexts.

Our study focused on analyzing a short and imbalanced dataset obtained by undersampling a larger dataset. We aimed to benchmark our proposed hybrid Q-CNN against CML, QML, and RL models.  In working with a short and imbalanced dataset, our hybrid model showcased its strength and outperformed the CML, QML, and RL models. It is essential to emphasize that our hybrid model's superior performance on this particular short and imbalanced dataset highlights its effectiveness in addressing the specific challenges associated with such data characteristics. This milestone underscores the practicality and utility of the hybrid Q-CNN in real-world scenarios where acquiring large datasets may be difficult, yet accurate predictions are crucial. Our findings have implications for domains with similar short and imbalanced datasets. The success of our proposed model indicates its practicality and usefulness in situations where obtaining extensive datasets is challenging, but accurate predictions are of paramount importance.

\section*{Methods}
\label{sec:methods}
\subsection*{Data Collection}
We used a benchmarking dataset called \href{https://www.kaggle.com/datasets/gowthammiryala/back-order-prediction-dataset}{\textit{``Can you predict product backorder?"}} obtained from the Kaggle data repository to conduct extensive experiments on our proposed hybrid Q–CNN–based prediction model. The data was gathered from the Kaggle repository. There are many orders for various products included in the dataset. A total of 22 features characterize the eight-week trajectory of each order, and a target binary feature denotes if the corresponding product is a backorder or not. Table \ref{table:features} summarizes the features.

One significant challenge in working with this dataset is the inherent class imbalance related to product backorders. Notably, backorder instances are relatively uncommon in real-world SC scenarios. In this dataset, out of a total of 1,929,935 orders, only 13,981 orders (approximately 0.72\%) are classified as delayed or experiencing a backorder. Conversely, a vast majority of 1,915,954 orders (about 99.28\%) are categorized as negative cases, indicating that the products were not subject to backorders. Figure \ref{fig3} shows the dataset's class distribution. This substantial class imbalance poses a notable challenge for predictive modeling. With an imbalance ratio of approximately 1:137, the dataset is heavily skewed toward the negative class, making it highly unbalanced. This imbalance can significantly impact the performance of ML models, as they may tend to favor the majority class, resulting in suboptimal predictive accuracy for the minority class. 

To address this challenge and ensure a rigorous evaluation of our proposed model, we adopted a stratified k-fold cross-validation approach with five splits, ensuring that the class distribution is maintained in each fold's training and testing sets. Additionally, we chose not to shuffle the dataset during this process, preserving the integrity of the class distribution.

\begin{figure*}[!ht]
  \centering
  \includegraphics[width=0.9\textwidth, height=0.75\textwidth]{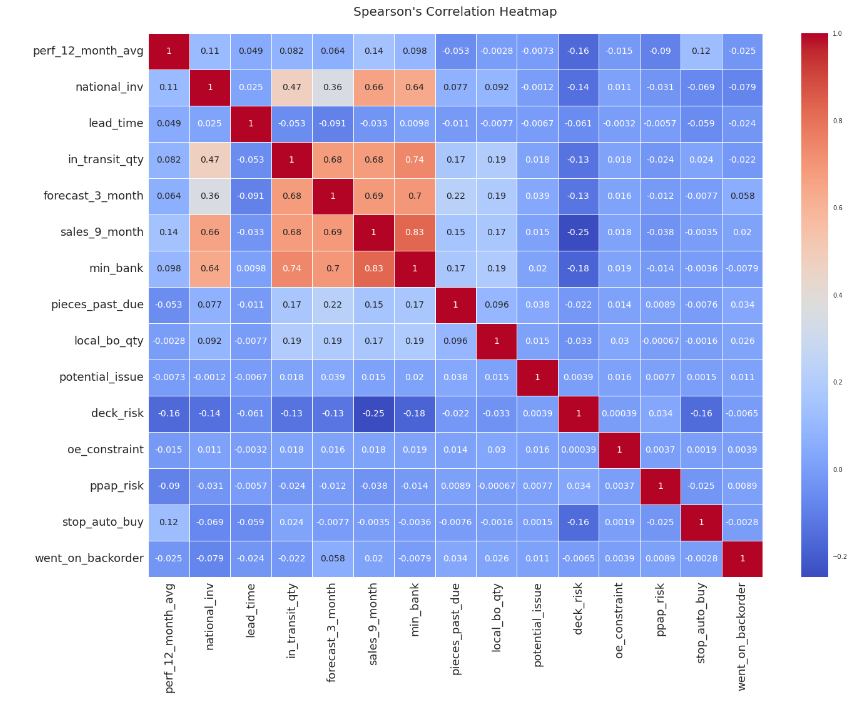}
  \caption{Spearman correlation heatmap to analyze the relationship between the target feature 'went\_on\_backorder' and 14 preprocessed input features of the SCBP dataset.}
  \label{fig2}
\end{figure*}

\begin{figure}[!ht]
  \centering
  \includegraphics[width=0.5\textwidth, height=0.32\textwidth]{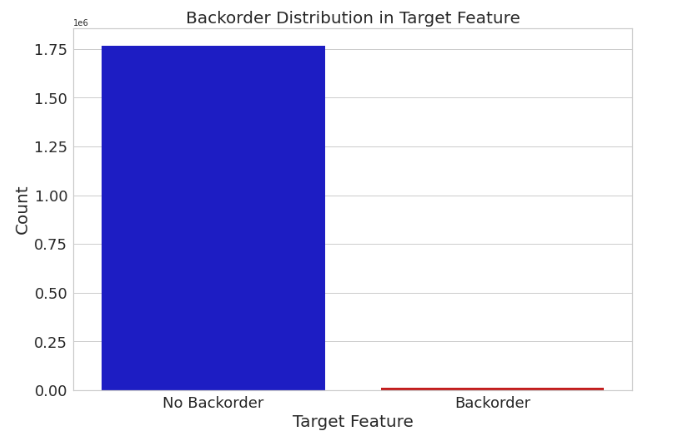}
  \caption{Class distribution of the imbalanced dataset used in our study.}
  \label{fig3}
\end{figure}


\begin{table}[!ht]
\centering
\resizebox{\linewidth}{!}{%
\begin{tabular}{|l|c|l|} 
\hline
\textbf{Features}                                                  & \textbf{Notation} & \textbf{Description}                                                                                                                                         \\ 
\hline
sku (Stock Keeping Unit)                                           & x0                & A distinctive identifier for each instance in the dataset.                                                                                                             \\ 
\hline
national\_inv                                                      & x1                & Current level of inventory of the product.                                                                                                                  \\ 
\hline
lead time                                                          & x2                & Time taken for a shipment to be delivered from its starting point to the final destination.                                                                  \\ 
\hline
in\_transit\_qty                                                   & x3                & This quantity is calculated based on the most recent picking slip or the cumulative\\& & quantity, and it represents the amount of product currently in transit.  \\ 
\hline
forecast\_3\_months, forecast\_6\_months, forecast\_9\_months      & x4, x5, x6        & Sales forecasts for the product over the following three, six, and nine months, respectively.                                                                              \\ 
\hline
sales\_1\_month, Sales\_3\_month, sales\_6\_month,
sales\_9\_month
& x7, x8, x9, x10   & Sales quantity of the product in the last one, three, six, and nine months, respectively.                                                                                  \\ 
\hline
min\_bank                                                          & x11               & Minimum recommended stocking level for the product.                                                                                                         \\ 
\hline
potential\_issue                                                   & x12               & Any problems or issues associated with the product or its parts.                                                                                             \\ 
\hline
pieces\_past\_due                                                  & x13               & Quantity of overdue parts for the product, if any.                                                                                                           \\ 
\hline
perf\_6\_months\_avg, perf\_12\_months\_avg                        & x14, x15          & Average performance of the product over the past six months and twelve months, respectively.                                                                       \\ 
\hline
local\_bo\_qty                                                     & x16               & Amount of stock orders that are currently overdue.                                                                                                           \\ 
\hline
ppap\_risk, deck\_risk, stop\_auto\_buy, oe\_constraint,  rev\_stop & x17 -- x21         & Binary flags (yes or no) associated with specific risks \\& & or constraints related to the products.                                                              \\ 
\hline
went\_to\_backorder                                                & y                 & Target variable by which the status of the product's backorder is indicated.                                                                                   \\
\hline
\end{tabular}
}
\caption{Descriptions of the dataset features of a particular product order.}
\label{table:features}
\end{table}

\subsection*{Data Preprocessing}
Data preprocessing is a crucial step in enhancing the performance of ML models. Our preprocessing approach initially focused on identifying and addressing irrelevant data points. For instance, we observed that variables like perf\_6\_month\_avg and perf\_12\_month\_avg contained negative values deemed inconsistent and removed. We encountered features that included the symbol '?' indicating missing values, which were also eliminated. Furthermore, we transformed categorical features into binary numerical representations to facilitate analysis, separating them from the original dataset for subsequent processing and analysis. For instance, the value of certain features containing either `yes' or `no' were converted into binary 1 and 0, respectively.

We tried seven different combinations of preprocessing steps and tested LR on each preprocessed data to evaluate and choose the best preprocessing step for our model development. Seven alternative techniques are as follows:
\begin{enumerate} 

\item\textit{IFLOF:} We removed anomalies from the dataset using the Isolation Forest Local Outlier Factor (IFLOF), which is a method that combines the Isolation Forest algorithm with the Local Outlier Factor algorithm. IFLOF identifies outliers by constructing an ensemble of isolation trees and measuring the local outlier factor for each data point. It provides a measure of abnormality for each instance in the dataset.

\item\textit{IFLOF+VIF:} This preprocessing step combines IFLOF outlier detection with the VIF. VIF is a measure of multicollinearity, which assesses the correlation between predictor variables in a regression model. Applying IFLOF to identify outliers and then using VIF to identify highly correlated variables helps address outlier detection and multicollinearity issues.

\item\textit{IQR+VIF:} We applied the Interquartile Range (IQR), a statistical dispersion measure, to identify outliers and then applied the VIF to detect and remove multicollinearity.

\item\textit{VIF:} We only applied VIF in this method without using any log transformation, standard scaling, or anomaly detection algorithm.

\item\textit{No log transform+VIF:} This preprocessing step involves applying VIF to the dataset without performing a log transform on the variables. This method allows for the detection of multicollinearity without the influence of log transformations.

\item\textit{RobScaler+VIF:} In this alternative, we tried RobScaler, a method used for robust feature scaling to the dataset, before using VIF to detect multicollinearity. RobScaler is particularly useful when dealing with data that contains outliers, as it scales the features by removing the median and scaling them according to the interquartile range.

\item\textit{Log transform+StandardScaler+VIF:} This proposed preprocessing step involves three stages. First, a log transform is applied to the variables, which can help normalize skewed data and handle nonlinear relationships. We removed the infinity values from the resulting data. Then, the StandardScaler is used for feature scaling, which ensures that all variables have a mean and standard deviation of 0 and 1, respectively. Finally, the VIF with threshold = 5 is applied to detect multicollinearity in the transformed and scaled dataset. This method aims to handle skewness, standardize the data, and identify multicollinearity simultaneously. We selected a subset of 14 features (x1–x4, x10–x13, x15, x16, x17–x20) from the dataset in this technique. We made this selection by excluding the remaining features due to the existence of multicollinearity among them.
\end{enumerate}

To choose the best preprocessing method, we undersampled the dataset to make it balanced, maintaining the majority-to-minority ratio of 3:1 using the NearMiss algorithm. Compared to the other approaches, Log transform+StandardScaler+VIF produces the best ROC-AUC of 66\% by the LR model (see Table \ref{table:preprocessing}). LR was chosen specifically to evaluate the performance of different preprocessing methods because it is a widely used and well-established classification algorithm. By applying various preprocessing methods and evaluating their effects on LR's performance, we gain valuable insights into which techniques are most effective in improving the predictive capabilities of LR.

After selecting the best preprocessing technique, we finally balanced and shortened the preprocessed dataset using the NearMiss algorithm, which was further fed as the input data for all the models used in this study. The input training data has 1000 samples having a 1:1 majority-to-minority class ratio. The test data was intentionally made imbalanced using the undersampling majority-to-minority ratio of 3:1, having 267 samples, among which 67 went backorder and the rest did not.

\begin{table}
\centering
\resizebox{\linewidth}{!}{%
\begin{tabular}{|l|l|l|l|l|l|l|l|l|} 
\hline
\multicolumn{2}{|l|}{\multirow{2}{*}{\textbf{Performance Metrics}}} & \multicolumn{7}{l|}{\textbf{Preprocessing Techniques}}                                                                                                                                 \\ 
\cline{3-9}
\multicolumn{2}{|l|}{}                                                          & \textbf{IFLOF} & \textbf{IFLOF+VIF} & \textbf{IQR+VIF} & \textbf{VIF} & \textbf{No log transform+VIF} & \textbf{RobScaler+VIF} & \textbf{Log transform+StandardScaler+VIF (Proposed)}  \\ 
\hline
\multirow{3}{*}{Not Backorder (0)} & Precision                                  & 99\%           & 99\%               & 99\%             & 100\%        & 100\%                         & 99\%                   & 100\%                                                 \\ 
\cline{2-9}
                                   & Recall                                     & 17\%           & 19\%               & 21\%             & 40\%         & 58\%                          & 100\%                  & 59\%                                                  \\ 
\cline{2-9}
                                   & F1-score                                   & 29\%           & 31\%               & 35\%             & 58\%         & 73\%                          & 100\%                  & 74\%                                                  \\ 
\hline
\multirow{3}{*}{Backorder (1)}     & Precision                                  & 1\%            & 1\%                & 1\%              & 1\%          & 1\%                           & 0\%                    & 1\%                                                   \\ 
\cline{2-9}
                                   & Recall                                     & 84\%           & 82\%               & 79\%             & 88\%         & 62\%                          & 0\%                    & 72\%                                                  \\ 
\cline{2-9}
                                   & F1-score                                   & 1\%            & 1\%                & 2\%              & 2\%          & 2\%                           & 0\%                    & 2\%                                                   \\ 
\hline
\multicolumn{2}{|l|}{Accuracy}                                                  & 17\%        & 19\%            & 22\%          & 41\%      & 58\%                       & 99\%                & 60\%                                               \\ 
\hline
\multicolumn{2}{|l|}{ROC AUC}                                                   & 50\%        & 50\%            & 50\%          & 64\%      & 60\%                       & 50\%                & \textbf{66\%}                                               \\ 
\hline
\multicolumn{2}{|l|}{Micro average precision}                                   & 50\%        & 50\%            & 50\%          & 50\%      & 50\%                       & 50\%                & 50\%                                               \\ 
\hline
\multicolumn{2}{|l|}{Micro average recall}                                      & 50\%        & 50\%            & 50\%          & 50\%      & 60\%                       & 50\%                & 66\%                                               \\
\hline
\end{tabular}
}

\caption{Performance evaluation of LR model on the undersampled data with different preprocessing techniques compared in this study}
\label{table:preprocessing}
\end{table}

\subsection*{Classical Models}
We implemented 8 CML models using the scikit-learn \cite{scikit} library, which provides a comprehensive set of tools for ML tasks in Python. Additionally, the parallel-computing library Dask \cite{dask} was utilized to enhance the efficiency and scalability of these algorithms. It enabled the distribution and execution of computations across multiple processors or machines, allowing for faster processing. We performed hyperparameter tuning using GridSearchCV with a 3-fold cross-validation to identify the optimal hyperparameters for each CML, as shown in Table \ref{gridsearch}.

\begin{table}
\centering

\small
\begin{tabular}{|l|l|} 
\hline
\textbf{Models}                     & \textbf{Best hyperparameters}                       \\ 
\hline
\multirow{22}{*}{\textbf{Catboost}} & boost\_from\_average = False                        \\ 
\cline{2-2}
                                    & boosting\_type = 'Plain'                            \\ 
\cline{2-2}
                                    & border\_count = 254                                 \\ 
\cline{2-2}
                                    & depth = 20                                          \\ 
\cline{2-2}
                                    & devices = '0:1'                                     \\ 
\cline{2-2}
                                    & early\_stopping\_rounds = 500                       \\ 
\cline{2-2}
                                    & eval\_metric = 'AUC'                                \\ 
\cline{2-2}
                                    & feature\_border\_type = 'GreedyLogSum'              \\ 
\cline{2-2}
                                    & grow\_policy = 'Lossguide'                          \\ 
\cline{2-2}
                                    & leaf\_estimation\_backtracking = 'Any Improvement'  \\ 
\cline{2-2}
                                    & learning\_rate = 0.5                                \\ 
\cline{2-2}
                                    & loss\_function = 'Logloss'                          \\ 
\cline{2-2}
                                    & max\_leaves = 100                                   \\ 
\cline{2-2}
                                    & model\_size\_reg = 0.5                              \\ 
\cline{2-2}
                                    & posterior\_sampling = False                         \\ 
\cline{2-2}
                                    & random\_seed = 786                                  \\ 
\cline{2-2}
                                    & random\_strength = 1                                \\ 
\cline{2-2}
                                    & rsm = 1                                             \\ 
\cline{2-2}
                                    & scale\_pos\_weight = 3                              \\ 
\cline{2-2}
                                    & score\_function = 'cosine'                          \\ 
\cline{2-2}
                                    & sparse\_features\_conflict\_fraction = 0            \\ 
\cline{2-2}
                                    & task\_type = 'GPU'                                  \\ 
\hline
\multirow{5}{*}{\textbf{LGBM}}      & colsample\_bytree = 1.0                             \\ 
\cline{2-2}
                                    & learning\_rate = 0.01                               \\ 
\cline{2-2}
                                    & max\_depth = 5                                      \\ 
\cline{2-2}
                                    & n\_estimators = 500                                 \\ 
\cline{2-2}
                                    & subsample = 0.8                                     \\ 
\hline
\multirow{7}{*}{\textbf{RF}}        & bootstrap = True                                    \\ 
\cline{2-2}
                                    & criterion = 'gini'                                  \\ 
\cline{2-2}
                                    & max\_depth = 30                                     \\ 
\cline{2-2}
                                    & max\_features = 'auto'                              \\ 
\cline{2-2}
                                    & min\_samples\_leaf = 2                              \\ 
\cline{2-2}
                                    & min\_samples\_split = 2                             \\ 
\cline{2-2}
                                    & n\_estimators = 200                                 \\ 
\hline
\multirow{6}{*}{\textbf{XGBoost}}   & learning\_rate = 0.3                                \\ 
\cline{2-2}
                                    & max\_depth = 20                                     \\ 
\cline{2-2}
                                    & min\_child\_weight = 2                              \\ 
\cline{2-2}
                                    & n\_estimators = 100                                 \\ 
\cline{2-2}
                                    & scale\_pos\_weight = 0.5                            \\ 
\cline{2-2}
                                    & colsample\_bytree = 1                               \\ 
\hline
\multirow{2}{*}{\textbf{KNN}}       & algorithm = 'ball\_tree'                            \\ 
\cline{2-2}
                                    & n\_neighbors = 4                                    \\ 
\hline
\multirow{2}{*}{\textbf{SVM}}       & kernel = 'rbf'                                      \\ 
\cline{2-2}
                                    & C = 0.9                                             \\ 
\hline
\multirow{3}{*}{\textbf{DT}}        & min\_samples\_leaf = 6                              \\ 
\cline{2-2}
                                    & criterion = 'gini'                                  \\ 
\cline{2-2}
                                    & max\_depth = 3                                      \\
\hline
\end{tabular}

\caption{Best hyperparameters selected by GridSearchCV for the CML models}
\label{gridsearch}
\end{table}

The CML models used include Categorical Boosting (Catboost), Light Gradient Boosting Machine (LGBM), Random Forest (RF), Extreme Gradient Boosting (XGBoost), Artificial Neural Network (ANN), K-Nearest Neighbors (KNN), Support Vector Machines (SVM), and Decision Tree (DT). The classical ANN architecture employed in this study consists of an input layer with 14 neurons, followed by two dense layers with 14 and 10 neurons, respectively.

\subsection*{Stacked Ensemble Models}
Using qiskit \cite{qiskit} and qiskit\_machine\_learning modules, we explore the following classically-stacked quantum ensemble algorithm:
\begin{enumerate}
    \item The base classifiers are trained using the provided training data.
    \item The trained base classifiers are then used to make predictions on both train and test datasets.
    \item The output labels generated by the base classifiers on the training and testing data are appended as additional features to the original training and testing datasets.
    \item Next, the meta-classifier is trained using the updated train data, and its performance is evaluated on the updated testing data to obtain the final prediction values.
\end{enumerate}

\subsubsection*{QSVM+LGBM+LR}
We initialize two base classifiers, namely Quantum Support Vector Machine (QSVM) and LGBM. For the QSVM classifier, we utilize the ZZFeatureMap to calculate the kernel matrix. The computation of the kernel matrix is performed using the following equation:
\begin{equation}
  K(\vec{x_i},\vec{x_j}) = K_{ij} = \vert\langle\phi^\dagger(\vec{x_j})\vert\phi(\vec{x_i})\rangle\vert^2
\end{equation}

where \(x_i, x_j \in X\) (training dataset), and \(\phi\) represents the feature map. To simulate the results of the quantum computer, we employ the state vector simulator, which can be substituted with a backend for hardware results. We consider two base classifiers, with the second one being LGBM. For the ensemble construction, we utilize LR as the meta-classifier, which combines the predictions of the two base classifiers.

\subsubsection*{VQC+QSVM}

We used the ZZFeatureMap to define the feature map for the Variational Quantum Classifier (VQC) as a base classifier and QSVM as a meta-classifier. The input data was mapped to a higher-dimensional quantum space using this feature map. For the VQC, we chose the TwoLocal ansatz, which involved the use of $R_Y$ (Equation \ref{ry}) and $R_Z$ (Equation \ref{rz}) gates for the parameterized rotations and the $CZ$ (Equation \ref{cz}) gate for entanglement.  

\begin{equation}
   CZ = \begin{bmatrix}
1 & 0 & 0 & 0 \\
0 & 1 & 0 & 0 \\
0 & 0 & 1 & 0 \\
0 & 0 & 0 & -1 \\
\end{bmatrix} 
\label{cz}
\end{equation}

This ansatz was repeated for two iterations. Then COBYLA optimizer and a QuantumInstance with the statevector\_simulator backend were configured. The QSVM's kernel was initialized using the QuantumKernel, which employed the chosen feature map and a QuantumInstance with the statevector\_simulator backend.

\subsubsection*{VQC+LGBM}
We utilized the previously mentioned initialization techniques for both the VQC and LGBM models, employing them as the base and meta classifiers. These models were then integrated into a stacking ensemble framework, where the predictions of the base classifiers were combined and used as features for the meta-classifier LGBM.

\subsection*{Quantum Neural Network (QNN) Models}

We used Pennylane \cite{pennylane} dependencies for developing the QNN models. Pennylane is employed to simulate quantum circuits and conduct quantum experiments, facilitating the development of QC programs.

\subsubsection*{MERA-VQC}
Our scheme has an ansatz based on a tensor network named Multi-scale Entanglement Renormalization Ansatz (MERA). With only 16 variables designed, the amplitude embedded one layer for each tensor network. We initialized the device as the QC backend with 4 qubits using Pennylane’s QML library \cite{pennylane}. The entanglement structure for the MERA circuit was implemented using CNOT gates between the qubits 0 and 1, and two rotation gates, $R_Y$, were applied to each qubit using the specified weights. The number of block wires, parameters per block, and the total number of blocks parameterized the MERA quantum circuit. Then a quantum circuit was implemented using the defined MERA structure to process the training data. We defined a VQC classifier that utilized the previously constructed quantum circuit. The classifier took weights, bias, and classical data as inputs and produced predictions based on the output of the circuit. We implemented a square loss function to measure the difference between the predicted and true labels and an accuracy function to assess the model's performance. We defined a cost function that was used to optimize the weights and bias parameters to enable the model to learn from  the training dataset by quantifying the overall loss between the predictions and the true labels.

\subsubsection*{RY-CNOT-VQC}
RY-CNOT-VQC 6-layered classifier highlights the use of $R_Y$ and CNOT gates in the circuit structure, providing more detailed information about the model's architecture. We employed a 2-qubit simulator to translate classical vectors into quantum state amplitudes. The circuit was encoded using the method described by \cite{mottonen}. Also, following the work of \cite{nielsen2010}, we had to break down controlled Y-axis rotations into simpler circuits. The quantum state preparation process was defined using quantum gates such as $R_Y$ (rotation around the y-axis), controlled-NOT (CNOT), and $Pauli-X$ gates. The primary quantum circuit incorporates the state preparation process, applying multiple rotations layers based on the given weights. Then a function applies rotation gates on qubits 0 and 1 and performs a CNOT operation between them. The quantum circuit was evaluated on a test input by applying the state preparation process and estimating the expectation value of the $Pauli-Z$ operator on qubit 0.

\subsubsection*{Classical NN+Encoder+QNN}
As suggested in \cite{killoran2019}, this hybrid model is made up of a classical NN, an encoder circuit, and a QNN. There are two qumodes that make up the quantum circuit. Each vector entry was used as the parameter of available quantum gates to encode classical data into quantum states. Two 10-neuron hidden layers, each with an 'ELU' activation function and a 14-neuron output layer, comprise the classical NN. Then, 14 entries of the classical NN's output vectors are sent into squeezer, interferometers, displacement gates, and Kerr gates as input parameters. Kerr gates, Interferometer-1, interferometer-2, squeezers, and displacement gates were employed in the QNN's four-step sequence. Using the $Pauli-X$ gate's $\langle\phi_k|X|\phi_k\rangle$ expectation value, for the final state $|\phi_k\rangle$ of each qumode, a two-element vector [$\langle\phi_0|X|\phi_0\rangle$, $\langle\phi_1|X|\phi_1\rangle$] was constructed. The ROC value of this model is 71.09\%, and the closest threshold to optimal ROC is 54\%.

\subsection*{Deep Reinforcement Learning Model}
We used TensorFlow 2.3+ \cite{tensorflow} and TF Agents 0.6+ \cite{tf-agents} to implement Double Deep Q-Network (DDQN) \cite{lin2019}. By treating the classification problem as an Imbalanced Classification Markov Decision Process, DDQN predicts that the episode will end when the agent misclassifies a sample from the minority-class but not a majority-class sample. The training process involved 100,000 episodes, and a replay memory was used with a length matching the number of warmup steps. Mini-batch training was performed with a batch size of 32, and the Q-network was updated using 2,000 steps of data collected during each episode. The policy was updated every 500 steps, and a soft update strategy was employed with a blending factor of 1 to update the target Q-network every 800 steps. The model architecture consisted of three dense layers with 256 units and ReLU activation, followed by dropout layers with a rate of 0.2. The final layer directly outputted the Q-values. Adam optimization was applied with a learning rate of 0.00025, and future rewards were not discounted. The exploration rate decayed from 1.0 to min\_epsilon over $\frac{1}{10}th$ of the total episodes, and the minimum and final chance of choosing a random action was set to 0.5.

\subsection*{Proposed QAmplifyNet Model}
The provided Figure \ref{proposed_method} presents an overview of our proposed methodological framework. The first phase in the framework is gathering baseline information, which may include supplier efficiency, lead times, inventory levels, and product sales. Information on sales, supplier efficiency, inventory levels, and lead times for suppliers is gathered from a wide variety of data sources. These data are then combined and grouped into weekly time intervals for orders. The dataset is subsequently divided into training and testing sets. The collected data undergoes preprocessing using our suggested `Log transformation+Standard Scaling+VIF treatment' method to address the common anomalies found in manufacturing industrial sensor data. This involves eliminating inconsistent data points, managing null values, and scaling and normalizing the data within a specified range. We applied PCA on both the train and test datasets to prepare the input for our 2-qubit Amplitude Encoder, resulting in 4 features. This dimensionality choice aligns with the model's requirements, as it operates on $\log_2{4}$, which yields a 2-dimensional classical data input. The aggregated data is then prepared for predictive analytics, employing a hybrid Q-CNN named QAmplifyNet as the core component of the proposed framework. The classical layers process the input data, while the quantum layer performs quantum computations on the encoded data. This comprehensive framework enables us to effectively leverage the collected data and utilize the hybrid model for analysis and prediction purposes. 

\begin{figure*}[!ht]
    \centering
    \includegraphics[width=1\linewidth]{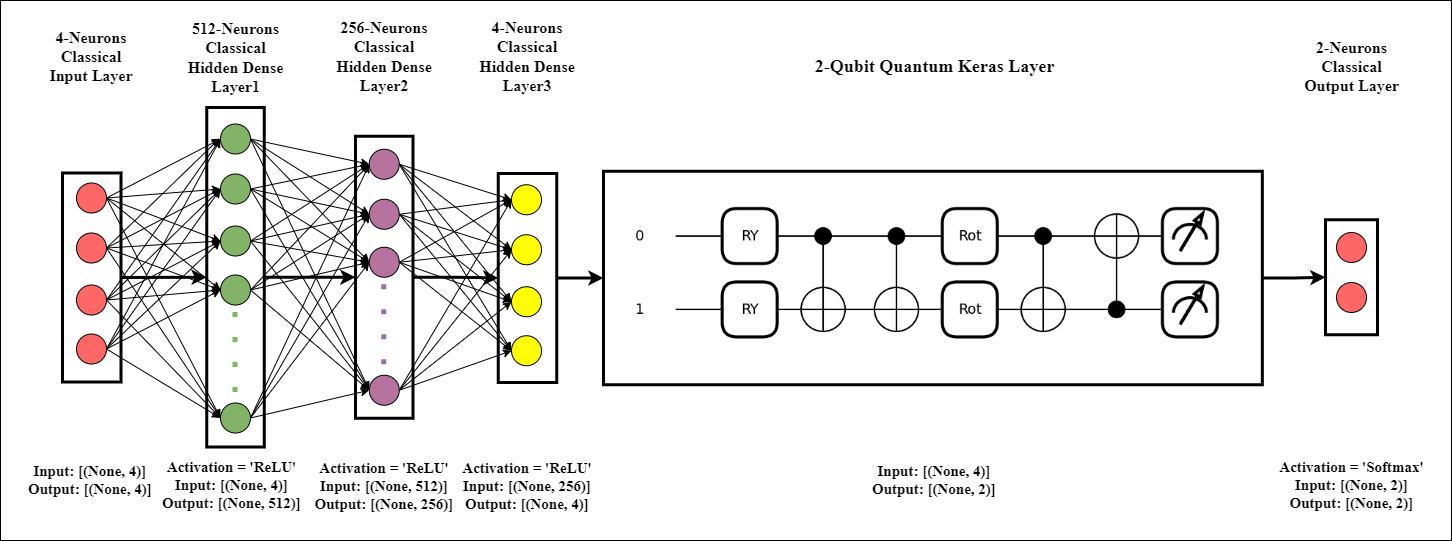}
    \caption{Model architecture of QAmplifyNet model.}
    \label{proposed_archi}
\end{figure*}

In our implementation, we leveraged the capabilities of PennyLane \cite{pennylane} to convert QNodes into Keras layers. This integration allowed us to combine these quantum layers with a diverse set of classical layers available in Keras, enabling the creation of genuinely hybrid models. Figure \ref{proposed_archi} explains the proposed architecture of QAmplifyNet, which consists of a Keras Sequential model consisting of an input layer, three classical hidden layers, one quantum layer, and a classical output layer. Here is an explanation of each layer:
\begin{enumerate}

\item\textit{Input Layer:} The input layer accepts inputs from 4 PC features and comprises 4 neurons.

\item\textit{Hidden Layer 1:} This dense layer has 512 units with a ReLU activation function. It receives input from the input layer (with a dimension of 4) and is set as non-trainable, serving the purpose of embedding the input data.

\item\textit{Hidden Layer 2:} The second dense layer has 256 units with a ReLU activation function. It receives input from the previous layer and is non-trainable.

\item\textit{Hidden Layer 3:} This dense layer has 4 units with a ReLU activation function. It takes input from the previous layer and is non-trainable. It passes 4-dimensional outputs to the next quantum layer.

\item\textit{QNN KerasLayer:} The next layer incorporates a 2-qubit quantum node (QNode) and weight shapes. It represents the quantum part of the hybrid model and takes input from the previous dense layer, which receives these four-dimensional classical data as inputs and converts them into four amplitudes, representing a quantum state of two qubits.

\item\textit{Output Layer:} The final output probabilities are generated via a softmax activation function in the output dense layer. There are just two possible classes that need to be classified; hence this layer only has 2 units. The softmax activation function can be characterized as follows:

\begin{equation}
    \sigma(\vec{\theta})_{i}=\frac{e^{\theta_{i}}}{\sum_{j=1}^{K} e^{\theta_{j}}}
\end{equation}

\begin{align*}
Where,\\
\sigma & = \text{softmax function} \\
\vec{\theta} & = \text{input vector} \\
e^{\theta_{i}} & = \text{standard exponential function for input vector} \\
K & = \text{class count in multi-classifier}
\end{align*}

\end{enumerate}

Using a learning rate of 0.01 and a loss function of 'binary\_crossentropy,' we employed the Adam optimizer.

In the QAmplifyNet mode, we have implemented distinct classical and quantum parts that work together to form the overall architecture. Let's delve into the details of each part:

\subsubsection*{Classical Part}
The classical part of the model primarily consists of classical layers that operate on classical data. In our specific implementation, we have used classical dense layers with various activation functions (e.g., ReLU) and configurations. These classical layers process the input data using classical computations, performing operations like linear transformations and nonlinear activations. Our model has three classical dense layers: Dense Layer 1, Dense Layer 2, and Dense Layer 3. These layers receive inputs from the previous layer and are set as non-trainable, as indicated by the `trainable=False` parameter. The classical part culminates in Dense Layer 4, which has two units and employs the Softmax activation function for generating the final output probabilities.

\subsubsection*{Quantum Part}
The quantum part of the model is integrated into the classical part using the `qml.qnn.KerasLayer' from PennyLane \cite{pennylane}. This part includes the QNode, which represents the quantum circuit, and weight shapes that define the structure of the quantum operations. In our implementation, the QNode is defined, which consists of  quantum operations from PennyLane's templates of `Amplitude Embedding' (AE) and `Strongly Entangling Layers' (SEL). 

\begin{figure}[!ht]
  \centering
  \includegraphics[width=0.9\textwidth, height=0.3\textwidth]{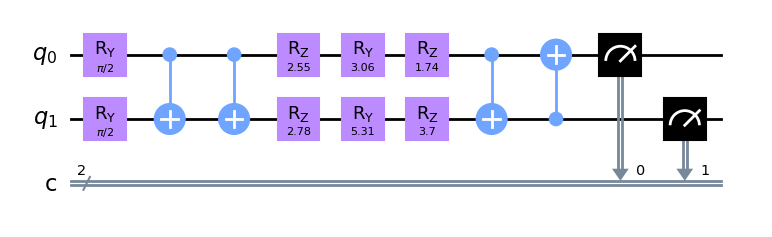}
  \caption{Quantum circuit representation of QAmplifyNet, featuring two qubits labeled "0" and "1". The circuit comprises variational layers utilizing the SEL approach for two qubits with a depth of one layer. Initial blue lines depict the embedding of features into the quantum state's amplitudes. Two $R_Y$-gates (see Equation \ref{ry}) introduce $\frac{\pi}{2}$ rotations to both qubits. Subsequently, two U3 rotation gates involving $R_Z$, $R_Y$, and $R_Z$ (see Equation \ref{rz}) single-qubit rotations are optimized during training. Blue CNOT (see Equation \ref{cnot}) entangling gates connect qubits 0 and 1, reinforcing their entanglement in a circular topology. The measurement layer includes two $Pauli-Z$ operators (Graphics generated using Pennylane-Qiskit).}
  \label{proposed_quantum_circuit}
\end{figure}

Classical data items must be embedded as quantum states on qubits for processing by a quantum computer due to the quantum nature of the computer's operation. In the circuit, the state preparation component, AE, is responsible for encoding classical data onto the two data qubits. The key advantage of AE is its ability to handle significantly large amounts of information with a relatively small number of qubits. With amplitude encoding, the number of amplitudes available is practically limitless, allowing for encoding a significant amount of data. Notably, the number of qubits required for encoding a given number of features follows a logarithmic relationship $(log_2(n))$, meaning that as the number of data features increases, only a logarithmic increase in the number of qubits is needed. This scalability enables encoding a vast amount of information with each additional qubit, making amplitude encoding a powerful approach for handling complex datasets in QC. The AE is composed of a parameterized quantum circuit comprising an embedding circuit and a variational circuit (see Figure \ref{proposed_quantum_circuit}). The embedding circuit incorporates an Amplitude Encoder, which is designed to encode a maximum of $2^n$ data features into the amplitudes of a quantum state consisting of $n$ qubits. Alternatively, a vector containing $N$ number of features can be encoded using [$\log_2 n$] qubits. The amplitudes of a quantum state $|\phi_x\rangle$ with $n$ qubits can be thought of as a representation of a normalized classical datapoint $x$ with $N$ dimensions, as 
\begin{equation}
    |\phi_x\rangle = \sum_{i=1}^{N} x_i|i\rangle
\end{equation}

In the given equation, where $N$ is equal to $2^n$, $x_i$ represents the $i$-th element of the variable $x$, and $|i\rangle$ refers to the $i$-th state in the computational basis. Nevertheless, $x_i$ can be a float or integer. The $x$ vector must be normalized according to the definition.

\begin{equation}
    \sum_{i=1} |x_i|^2 = 1
\end{equation}

If the number of features to encode is not a power of 2, the remaining amplitudes can be filled with constant values. The AE technique transforms the 4 features obtained from the classical component into the amplitudes of a quantum state with 2 qubits.

\begin{equation}
R_Y(\phi) = \begin{bmatrix}
\cos(\phi/2) & -\sin(\phi/2) \\
\sin(\phi/2) & \cos(\phi/2)
\end{bmatrix}
\label{ry}
\end{equation}

\begin{equation}
R_Z(\psi) = \begin{bmatrix}
e^{-i\psi/2} & 0 \\
0 & e^{i\psi/2}
\end{bmatrix}
\label{rz}
\end{equation}

\begin{equation}
CNOT = \begin{bmatrix}
1 & 0 & 0 & 0 \\
0 & 1 & 0 & 0 \\
0 & 0 & 0 & 1 \\
0 & 0 & 1 & 0 \\
\end{bmatrix}
\label{cnot}
\end{equation}

In the variational stage, the number of SEL, $L$ is variable. SEL consists of generic trainable rotational gates $Rot(\alpha_i,\beta_i,\gamma_i)$ implemented on qubits 0 and 1, and then a set of CNOT gates are used to connect adjacent qubit pairs, with the last qubit being regarded as a neighbor of the first. The number of SEL for an $n$-qubit circuit can be modified to tune the complexity of this circuit. The model has precisely $3\times{n}\times{L}$ number of trainable parameters. The SEL utilizes a circuit-centric approach in its design. In this approach, each individual qubit, denoted as $G$, is represented by a $2\times2$ unitary matrix, as shown in Equation \ref{eq:strongly_entangling}, where $\theta,\phi,\psi \in [0,\pi]$.

\begin{equation}
G(\theta,\phi,\psi) = \begin{bmatrix}
e^{i\phi}\cos(\theta) & e^{i\psi}\sin(\theta) \\
-e^{-i\psi}\sin(\theta) & e^{-i\phi}\cos(\theta)
\end{bmatrix}
\label{eq:strongly_entangling}
\end{equation}

Due to the lack of support for the "reversible" differentiation method in SEL, PennyLane \cite{pennylane} automatically chooses the most suitable differentiation method available. The state of the two qubits can be measured using the $Pauli-Z$ operator. Upon measurement, the qubits will collapse to a specific state. The matrix representation of the $Pauli-Z$ operator is illustrated in Equation (\ref{eq:pauli_z}).

\begin{equation}
    \sigma_z = \begin{bmatrix}
1 & 0 \\
0 & -1 \\
\end{bmatrix}
\label{eq:pauli_z}
\end{equation}

The measurement of the first qubit's $Pauli-Z$  operator is denoted as $\langle\sigma_z^0\rangle\in[-1,+1]$. This expectation value is subsequently utilized to determine the probabilities involved $P_{not backorder}$ and $P_{backorder}$ of being "not backorder" or "backorder" state, respectively:
\begin{equation}
    P_{not backorder} = \frac{1}{2}(\langle\sigma_z^0\rangle+1)
\end{equation}
\begin{equation}
    P_{backorder} = \frac{1}{2}(1-\langle\sigma_z^0\rangle) = 1 - P_{not backorder}
\end{equation}

These quantum operations help encode the input features into quantum states and perform quantum computations. The QNode calculates the $Pauli-Z$ operators' expectation values for each quantum circuit qubit. The QNode's output is then input for the subsequent classical layers.

\subsubsection*{Combining Classical and Quantum Parts}
The classical and quantum parts of the model are seamlessly integrated within the `Sequential' framework of Keras. The classical layers process the data up to a certain point, and then the output is fed into the quantum layer (KerasLayer), which incorporates the QNode. The challenge lies in the necessity of the 4-neuron dense layer preceding the 2-qubit quantum layer, as it serves to pass the essential 4-dimensional input features to the quantum layer, rendering its removal or replacement unfeasible. The Adam optimizer is utilized to train the parameters of the model, which include the weights and biases. The model is optimized during training based on the binary cross-entropy loss function. The training steps involve iteratively updating the parameters to improve the model's performance and accuracy. We also enabled the 'EarlyStopping' mechanism during the training process, ensuring that the model stops when the desired metric stops improving, which helped prevent overfitting and saved training time. We ensured that the model's primary goal was to generalize effectively to new, unseen data, which was confirmed during testing. Although the validation loss remained higher, the model achieved a good balance between fitting the training data and generalizing to new instances. The use of 'EarlyStopping' added robustness to our convergence analysis, preventing overfitting during training. The training procedure took place on a Kaggle kernel environment equipped with 2 CPU cores, 13 GB RAM, and 2 Nvidia Tesla T4 GPUs, each of 15 GB. The model parameters were carefully selected through multiple trial runs to optimize accuracy. The training process concluded after 18 epochs using a batch size of 5.

 The loss curve of Figure \ref{all_metrics} indicates the model's ability to minimize errors during the training and validation phases. Loss curves aid in evaluating the model's learning progress, generalization capability, and potential for effective predictions on new instances.

\begin{figure*}[!ht]
  \centering
    \includegraphics[width=1\textwidth,height=0.32\textheight]{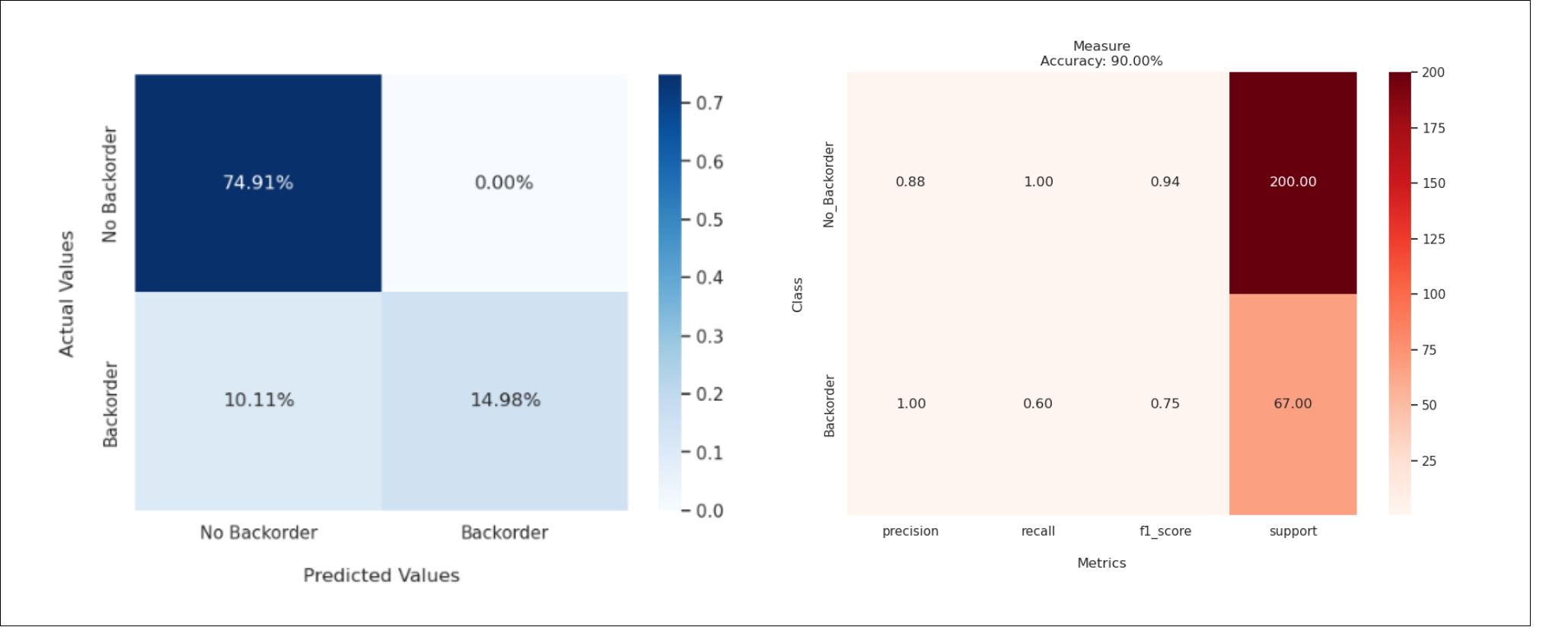}
  
  \caption{Confusion matrix and classification reports (from left to right) for QAmplifyNet model.}
  \label{proposed_cm_cr}
\end{figure*}

\begin{figure*}[!ht]
  \centering
  \includegraphics[width=0.95\textwidth,height=0.25\textheight]{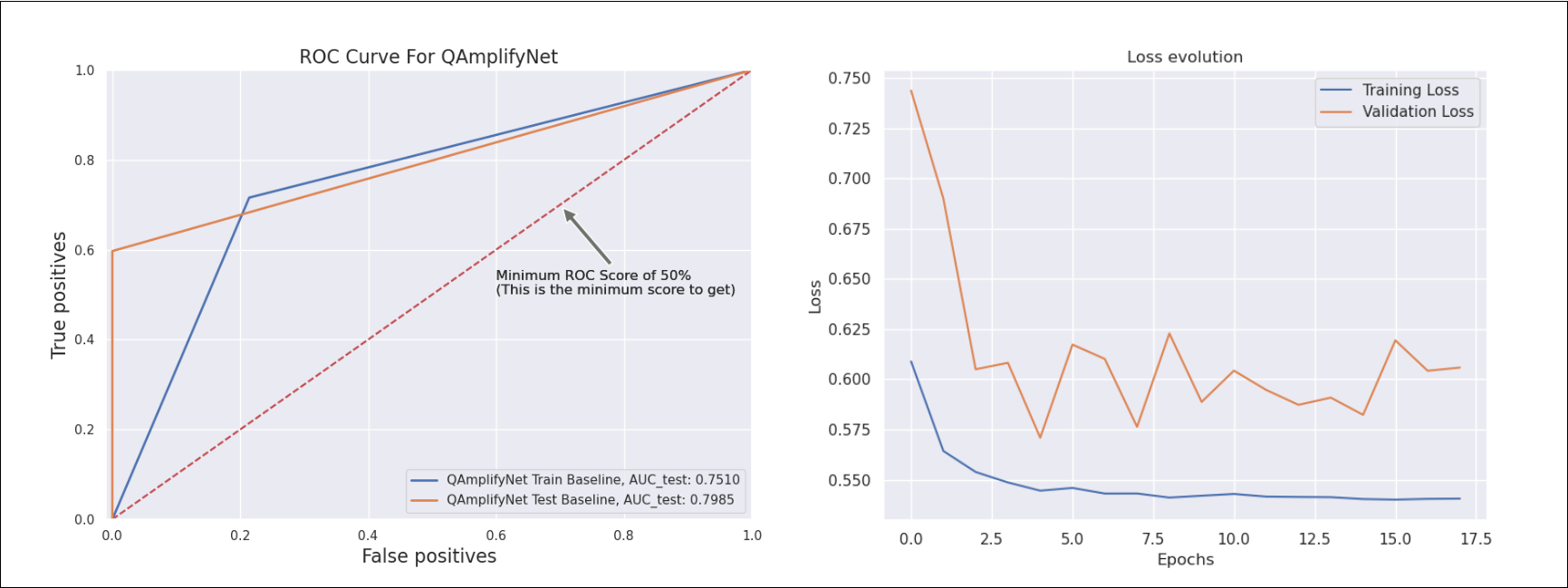}
    \label{proposed_accuracy_loss:subfig1}

  \caption{From left to right, the curves represent the ROC and Loss evolution vs. Epochs of the QAmplifyNet model. These curves provide insights into the model's ability to classify backorder instances accurately and its overall predictive performance as training progresses.}
  \label{all_metrics}
\end{figure*}

\section*{Results}
\label{sec:results}
\subsection*{Evaluation Metrics}
In order to assess the effectiveness of the predictive models employed in this study, various performance metrics have been utilized. In the context of SCBP, True Positive (TP) refers to the number of correctly classified instances of backorder occurrences, while True Negative (TN) represents the number of correctly classified instances of non-backorder occurrences. False Positive (FP) indicates the number of backorder instances mistakenly classified, and False Negative (FN) signifies the number of misclassified non-backorder instances. Both FP and FN are significant as higher FN results in missed opportunities with potential customers, leading to increased opportunity costs. On the other hand, higher FP leads to increased inventory holding costs and a greater risk of product obsolescence due to the long-term accumulation of unnecessary inventory. Here are the definitions and equations for the performance metrics used to evaluate our models:

\subsubsection*{Accuracy}
Accuracy is a metric that evaluates the overall accuracy of predictions by determining the ratio of correct predictions to the total number of predictions made.

\begin{equation}
    \text{{Accuracy}} = \frac{{TN + TP}}{{TN + TP + FN + FP}}
\end{equation}

\subsubsection*{Precision}
Precision is a metric that evaluates the accuracy of positive predictions made by a model. It represents the proportion of correctly identified positive instances out of all instances that were predicted as positive. This metric helps assess the model's capability to minimize false positive predictions, providing insights into its precision and reliability in identifying positive cases accurately.

\begin{equation}
    \text{{Precision}} = \frac{{TP}}{{FP + TP}}
\end{equation}

\subsubsection*{Recall (True Positive Rate)}
Recall is a metric that quantifies the model's effectiveness in correctly identifying positive instances among all the actual positive instances. It provides insight into how well the model can detect and capture the positive cases in the dataset.

\begin{equation}
    \text{{Recall}} = \frac{{TP}}{{TP + FN}}
\end{equation}

\subsubsection*{F1-measure}
The F1-measure is a metric that combines precision and recall into a single value, giving equal importance to both. It serves as a balanced measure that is particularly beneficial when dealing with imbalanced datasets. By considering both precision and recall, the F1 measure provides a comprehensive evaluation of a model's performance, considering both the ability to correctly identify positive instances (precision) and capture all positive instances (recall). This makes it a valuable metric in uneven class distribution scenarios like the SCBP dataset, as it offers a balanced assessment of the model's effectiveness.

\begin{equation}
    \text{{F1-measure}} = 2 \times \frac{{\text{{Precision}} \times \text{{Recall}}}}{{\text{{Precision}} + \text{{Recall}}}}
\end{equation}

\subsubsection*{Specificity (True Negative Rate)}
Specificity is a metric that quantifies the accuracy of a model in correctly identifying negative instances among all the actual negative instances. It provides insight into the model's capability to detect and classify negative instances accurately.

\begin{equation}
    \text{{Specificity}} = \frac{{TN}}{{TN + FP}}
\end{equation}

\subsubsection*{Gmean}
The Gmean is a metric represented by the equation \ref{gmean} that aims to achieve a balance between maximizing the TP and TN. It takes into account both TP and TN while minimizing the adverse effects caused by imbalanced class distributions. It is crucial to acknowledge that the Gmean metric does not offer insights into the specific contributions made by each class towards the overall index. Consequently, various combinations of TN and TP can result in identical Gmean values.

\begin{equation}
    \text{{Gmean}} = \sqrt{{\text{{TP}} \times \text{{TN}}}}
\label{gmean}
\end{equation}

\subsubsection*{IBA}
IBA is a measure to estimate the performance of binary classifiers on imbalanced datasets using the following equation: 

\begin{equation}
    \text{{IBA}} = \text{(Gmean)}^2 \times (1 + \text{Dominance})
\end{equation}

Here, $Dominance$ refers to the absolute difference between TP and TN, which is utilized to gauge the relationship between these two measures. By substituting $Dominance$ and $Gmean$ into the equation, we can gain valuable insights into how the IBA balances the trade-off between $Dominance$ and the $Gmean$.

\subsubsection*{AUC-ROC index}
The AUC is a metric that evaluates the overall performance of a model by considering its ability to differentiate between positive and negative instances at various classification thresholds. It is represented graphically as a ROC curve. The AUC serves as an indicator of the model's discriminative power and its capacity to classify different instances accurately.

These performance metrics are relevant to the SCBP problem as they provide insights into the model's accuracy, precision, recall, and ability to handle imbalanced datasets. They help assess the model's effectiveness in correctly identifying backorder and non-backorder instances.

\subsection*{Results and Analysis}
The results presented in Table \ref{table:eval_metric} demonstrate the performance comparison of different algorithms for the task at hand. Our proposed QAmplifyNet algorithm achieves the highest accuracy score of 90\%, outperforming all the other models used in this study. Among the QNN models, MERA 4-layered, Classical NN+Encoder+QNN, and RY-CNOT 6-layered exhibit respectable accuracy scores of 78\%, 77\%, and 75\%, respectively. Nevertheless, when choosing an ML algorithm, it is vital to take into account factors beyond accuracy as the sole criterion. Considerations such as the algorithm's ability to generalize well to unseen data, its interpretability in providing understandable insights, and its computational efficiency should also be taken into consideration.

In our scenario, we have two classes: '0' represents "Not Backorder" and '1' represents "Backorder." While different models demonstrate better performance in either precision or recall, it is essential to consider both measures by assessing the F1-score. QAmplifyNet achieves the best macro-average F1-score of 84\%, with 94\% for predicting class 0 and 75\% for predicting class 1. Given the imbalanced nature of the dataset, we employed the `imblearn' module from scikit-learn to gain insights into specificity, Gmean, and IBA values. QAmplifyNet yields the highest Gmean (77\%) and IBA scores (62\% for class 0 and 57\% for class 1), outperforming the other models. Furthermore, QAmplifyNet achieved an AUC-ROC value of 79.85\%, indicating that the model exhibits stronger discriminatory power than the other models. The AUC-ROC analysis allows us to assess the model's overall ability to rank instances correctly and provides insights into their predictive capabilities.

QAmplifyNet achieves the highest macro-average precision and recall scores of 94\% and 80\%, respectively. Regarding class 0, QAmplifyNet achieves a precision of 88\% (see Figure \ref{proposed_cm_cr}), indicating that 88\% of instances predicted as class 0 are correctly classified. The recall of 100\% signifies that the model successfully identifies all true class 0 instances. The specificity of 60\% suggests that the model accurately identifies 60\% of the true class 1 instances as class 1. Concerning class 1, the precision of 100\% reveals that all instances predicted as class 1 are classified correctly. Nevertheless, with a recall of 60\%, it signifies that the model only manages to identify 60\% of the actual instances belonging to class 1. On the other hand, a specificity of 100\% implies that all instances belonging to class 0 are accurately classified as class 0. QAmplifyNet demonstrated significant outperformance compared to the other models, achieving a macro-average specificity of 80\%. This indicates that QAmplifyNet excelled in correctly identifying the negative instances, surpassing the performance of the other models in terms of distinguishing non-backorder cases accurately.

Notably, QAmplifyNet consistently demonstrates superior performance across all evaluation metrics: accuracy, AUC-ROC, precision, recall, F1-score, specificity, Gmean, and IBA (macro-average 59.50\%). In contrast, other models exhibit inconsistent performance across some of the metrics.

The comparison of confusion matrix components, namely TP, TN, FN, and FP, for various models in SCBP is depicted in the provided Figure \ref{barplot:tp,tn.fp.fn}. Upon analyzing the results, it becomes evident that QAmplifyNet outperforms other models in terms of predictive performance. QAmplifyNet achieved a TP rate of 14.98\% and a TN rate of 74.91\%, demonstrating its ability to classify positive and negative instances accurately. Significantly, it achieved a notable 0\% FP rate, signifying a complete absence of incorrect predictions labeling non-backorder instances as backorders. Furthermore, QAmplifyNet exhibited a relatively low FN rate of 10.11\%, implying a minimal number of missed positive predictions. In contrast, several models displayed higher FP rates, erroneously identifying actual non-backorders as backorders. Similarly, other models demonstrated higher FN rates, misclassifying backorder cases. This achievement is particularly significant given the imbalanced nature of the SCBP problem. For instance, the MERA 4-Layered and RY-CNOT 6-Layered models achieved 0\% FP rates but at the expense of higher FN rates of 22.10\% and 25.09\%, respectively. Additionally, their TP rates (3\% and 0\%) were lower than that of QAmplifyNet.

\begin{figure*}[!ht]
  \centering  \includegraphics[width=1\textwidth,height=0.4\textheight]{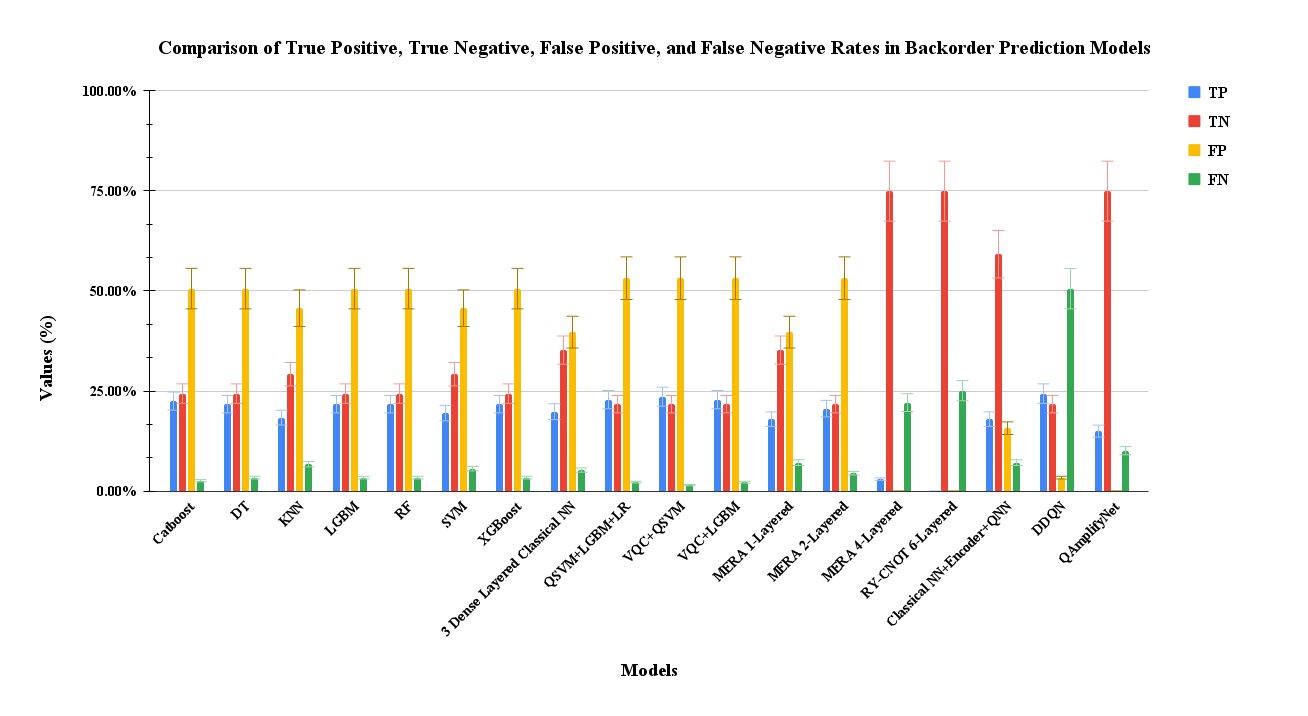}
  \caption{Bar plot illustrating the TP, TN, FP, and FN rates of various models used in this paper for SCBP. The obtained values are derived from the confusion matrices of each model, offering valuable information regarding their ability to accurately classify instances as either positive or negative.}
  \label{barplot:tp,tn.fp.fn}
\end{figure*}

Comparatively, the CML models exhibited significantly higher average FP rates (47.99\%) and relatively lower average TN rates (26.92\%). Similarly, the Stacked Ensemble models demonstrated FP rates of 53.18\% and TN rates of 21.72\%. Classical NN+Encoder+QNN had a 15.73\% higher FP rate and a 15.73\% lower TN rate compared to QAmplifyNet. It is worth noting that DDQN achieved a high TP rate of 24.34\% but at the cost of a substantial FP rate of 50.56\%. Conversely, QAmplifyNet achieved a competitive TN rate while maintaining a significantly lower FP rate of 0\%, underscoring its robustness in minimizing false positive predictions.

The comparison of QAmplifyNet with other models highlights its superiority in achieving a balanced trade-off between TP, TN, FP, and FN rates, resulting in more accurate SCBP with minimal FP and FN predictions. This substantiates QAmplifyNet's potential for enhancing the reliability and robustness of SCBP systems.

\begin{table}[!ht]
\centering
\resizebox{\linewidth}{!}{%
\begin{tabular}{|l|l|l|l|l|l|l|l|l|l|l|} 
\hline
\multirow{2}{*}{\textbf{Model Category}} & \multicolumn{2}{l|}{\multirow{2}{*}{\textbf{Models}}} & \multicolumn{8}{l|}{\textbf{Evaluation Metrics}}                                                                                                                                                  \\ 
\cline{4-11}
                                         & \multicolumn{2}{l|}{}                                 & \textbf{Precision} & \textbf{Recall} & \textbf{F1-score} & \textbf{Specificity} & \textbf{Gmean} & \textbf{IBA} & \textbf{Accuracy}                 & \textbf{ROC--AUC}                   \\ 
\hline
\multirow{20}{*}{CML Models}             & \multirow{2}{*}{Catboost}                 & 0         & 90\%            & 33\%         & 48\%           & 90\%              & 54\%                 & 27\%      & \multirow{2}{*}{47\%}          & \multirow{2}{*}{73.83\%}           \\ 
\cline{3-9}
                                         &                                           & 1         & 31\%            & 90\%         & 46\%           & 33\%              & 54\%                 & 31\%      &                                   &                                    \\ 
\cline{2-11}
                                         & \multirow{2}{*}{DT}                       & 0         & 88\%            & 33\%         & 47\%           & 87\%              & 53\%                 & 27\%      & \multirow{2}{*}{46\%}          & \multirow{2}{*}{68.07\%}           \\ 
\cline{3-9}
                                         &                                           & 1         & 30\%            & 87\%         & 45\%           & 33\%              & 53\%                 & 30\%      &                                   &                                    \\ 
\cline{2-11}
                                         & \multirow{2}{*}{KNN}                      & 0         & 81\%            & 39\%         & 53\%           & 73\%              & 53\%                 & 28\%      & \multirow{2}{*}{48\%}          & \multirow{2}{*}{55.03\%}           \\ 
\cline{3-9}
                                         &                                           & 1         & 29\%            & 73\%         & 41\%           & 39\%              & 53\%                 & 29\%      &                                   &                                    \\ 
\cline{2-11}
                                         & \multirow{2}{*}{LGBM}                     & 0         & 88\%            & 33\%         & 47\%           & 87\%              & 53\%                 & 27\%      & \multirow{2}{*}{46\%}          & \multirow{2}{*}{75.23\%}           \\ 
\cline{3-9}
                                         &                                           & 1         & 30\%            & 87\%         & 45\%           & 33\%              & 53\%                 & 30\%      &                                   &                                    \\ 
\cline{2-11}
                                         & \multirow{2}{*}{RF}                       & 0         & 88\%            & 33\%         & 47\%           & 87\%              & 53\%                 & 27\%      & \multirow{2}{*}{46\%}          & \multirow{2}{*}{73.96\%}           \\ 
\cline{3-9}
                                         &                                           & 1         & 30\%            & 87\%         & 45\%           & 33\%              & 53\%                 & 30\%      &                                   &                                    \\ 
\cline{2-11}
                                         & \multirow{2}{*}{SVM}                      & 0         & 84\%            & 39\%         & 53\%           & 78\%              & 55\%                 & 29\%      & \multirow{2}{*}{49\%}          & \multirow{2}{*}{63.74\%}           \\ 
\cline{3-9}
                                         &                                           & 1         & 30\%            & 78\%         & 43\%           & 39\%              & 55\%                 & 31\%      &                                   &                                    \\ 
\cline{2-11}
                                         & \multirow{2}{*}{XGBoost}                  & 0         & 88\%            & 33\%         & 47\%           & 87\%              & 53\%                 & 27\%      & \multirow{2}{*}{46\%}          & \multirow{2}{*}{71.90\%}           \\ 
\cline{3-9}
                                         &                                           & 1         & 30\%            & 87\%         & 45\%           & 33\%              & 53\%                 & 30\%      &                                   &                                    \\ 
\cline{2-11}
                                         & \multirow{2}{*}{3 Dense Layered NN}       & 0         & 87\%            & 47\%         & 61\%           & 79\%              & 61\%                 & 36\%      & \multirow{2}{*}{55\%}          & \multirow{2}{*}{72.52\%}           \\ 
\cline{3-9}
                                         &                                           & 1         & 33\%            & 79\%         & 47\%           & 47\%              & 61\%                 & 38\%      &                                   &                                    \\ 
\hline
\multirow{6}{*}{Stacked Ensemble Models} & \multirow{2}{*}{QSVM+LGBM+LR}             & 0         & 91\%            & 29\%         & 44\%           & 91\%              & 51\%                 & 25\%      & \multirow{2}{*}{45\%}          & \multirow{2}{*}{70.00\%}           \\ 
\cline{3-9}
                                         &                                           & 1         & 30\%            & 91\%         & 45\%           & 29\%              & 51\%                 & 28\%      &                                   &                                    \\ 
\cline{2-11}
                                         & \multirow{2}{*}{VQC+QSVM}                 & 0         & 94\%            & 29\%         & 44\%           & 94\%              & 52\%                 & 25\%      & \multirow{2}{*}{45\%}          & \multirow{2}{*}{62.00\%}           \\ 
\cline{3-9}
                                         &                                           & 1         & 31\%            & 94\%         & 46\%           & 29\%              & 52\%                 & 29\%      &                                   &                                    \\ 
\cline{2-11}
                                         & \multirow{2}{*}{VQC+LGBM}                 & 0         & 91\%            & 29\%         & 44\%           & 91\%              & 44\%                 & 25\%      & \multirow{2}{*}{45\%}          & \multirow{2}{*}{60.00\%}           \\ 
\cline{3-9}
                                         &                                           & 1         & 30\%            & 91\%         & 45\%           & 29\%              & 45\%                 & 28\%      &                                   &                                    \\ 
\hline
\multirow{10}{*}{QNN Models}             & \multirow{2}{*}{MERA 1-Layered}           & 0         & 83\%            & 47\%         & 60\%           & 72\%              & 58\%                 & 33\%      & \multirow{2}{*}{53\%}          & \multirow{2}{*}{59.32\%}           \\ 
\cline{3-9}
                                         &                                           & 1         & 31\%            & 72\%         & 43\%           & 47\%              & 58\%                 & 35\%      &                                   &                                    \\ 
\cline{2-11}
                                         & \multirow{2}{*}{MERA 2-Layered}           & 0         & 83\%            & 29\%         & 43\%           & 82\%              & 49\%                 & 23\%      & \multirow{2}{*}{42\%}          & \multirow{2}{*}{55.54\%}           \\ 
\cline{3-9}
                                         &                                           & 1         & 28\%            & 82\%         & 42\%           & 29\%              & 49\%                 & 25\%      &                                   &                                    \\ 
\cline{2-11}
                                         & \multirow{2}{*}{MERA 4-Layered}           & 0         & 77\%            & 100\%        & 87\%           & 12\%              & 35\%                 & 13\%      & \multirow{2}{*}{78\%}          & \multirow{2}{*}{55.97\%}           \\ 
\cline{3-9}
                                         &                                           & 1         & 100\%           & 12\%         & 21\%           & 100\%             & 35\%                 & 11\%      &                                   &                                    \\ 
\cline{2-11}
                                         & \multirow{2}{*}{RY-CNOT 6-Layered}        & 0         & 75\%            & 100\%        & 86\%           & 0\%               & 0\%                  & 0\%       & \multirow{2}{*}{75\%}          & \multirow{2}{*}{50.00\%}           \\ 
\cline{3-9}
                                         &                                           & 1         & 0\%             & 0\%          & 0\%            & 100\%             & 0\%                  & 0\%       &                                   &                                    \\ 
\cline{2-11}
                                         & \multirow{2}{*}{Classical NN+Encoder+QNN} & 0         & 80\%            & 89\%         & 84\%           & 79\%              & 75\%                 & --     & \multirow{2}{*}{77\%}          & \multirow{2}{*}{71.09\%}           \\ 
\cline{3-9}
                                         &                                           & 1         & 71\%            & 53\%         & 61\%           & 79\%              & 75\%                 & --     &                                   &                                    \\ 
\hline
\multirow{2}{*}{Deep RL Model}           & \multirow{2}{*}{DDQN}                     & 0         & 88\%            & 33\%         & 47\%           & 87\%              & 53\%                 & 27\%      & \multirow{2}{*}{46\%}          & \multirow{2}{*}{47.58\%}           \\ 
\cline{3-9}
                                         &                                           & 1         & 30\%            & 87\%         & 45\%           & 33\%              & 53\%                 & 30\%      &                                   &                                    \\ 
\hline
\multirow{2}{*}{\textbf{Proposed}}       & \multirow{2}{*}{\textbf{QAmplifyNet}}     & 0         & \textbf{88\%}            & \textbf{100\%}        & \textbf{94\%}  & \textbf{60\%}              & \textbf{77\%}                 & \textbf{62\%}      & \multirow{2}{*}{\textbf{90\%}} & \multirow{2}{*}{\textbf{79.85\%}}  \\ 
\cline{3-9}
                                         &                                           & 1         & \textbf{100\%}           & \textbf{60\%}         & \textbf{75\%}  & \textbf{100\%}             & \textbf{77\%}                 & \textbf{57\%}      &                                   &                                    \\
\hline
\end{tabular}
}
\caption{Performance comparisons of the models used in this study against QAmpliNet on short SCBP dataset}
\label{table:eval_metric}
\end{table}

\subsection*{XAI Interpretation using LIME and SHAP}
To gain insights into the interpretability of QAmplifyNet, we applied two popular XAI techniques: Local Interpretable Model-agnostic Explanations (LIME) and SHapley Additive exPlanations (SHAP) in Python programming language. By employing these methods, we were able to gain insights into the model's predictions and provide explanations by identifying the specific contributions of individual features.

\subsubsection*{LIME}
LIME is a local interpretability method that provides explanations for individual predictions by approximating the model's behavior around specific instances. By introducing perturbation into the input data and tracking how the hybrid model's predictions changed, we were able to utilize LIME to provide potential explanations for the model's behavior. This process allowed us to identify the most significant features and understand their influence on the model's decision-making. LIME achieves this by generating local surrogate models around a particular instance of interest. These surrogate models are simpler and more interpretable, such as linear or decision tree models, and capture the local behavior of the complex model. LIME examines the significance and influence of individual aspects on the model's decision-making process by perturbation of the input features and evaluating the ensuing changes in predictions. The equation of LIME can be expressed as follows:

\begin{equation}
    \gamma(x) = \arg\min_{h \in H} L(f, h, \pi_x) + \lambda(h)
\end{equation}

Here, the loss function $L$ quantifies the similarity between the original, sophisticated model $f$ and the interpretable model $h$. The family of interpretable models is denoted by $H$, while $\pi_x$ represents the closeness of the instances being evaluated to a specific instance $x$. The term $(h)$ indicates the significance or importance assigned to the model $h$, which can involve additional weighting or importance factors. LIME aims to find the interpretable model $h$ that minimizes the loss function and adequately captures the complex model $f$ behavior while considering the proximity and criticality aspects.

\begin{figure*}[!ht]
  \centering
  \hfill
  \subfigure[LIME explanation for a single instance including predicted probabilities of each class]{
   \includegraphics[width=1\textwidth,height=0.08\textheight]{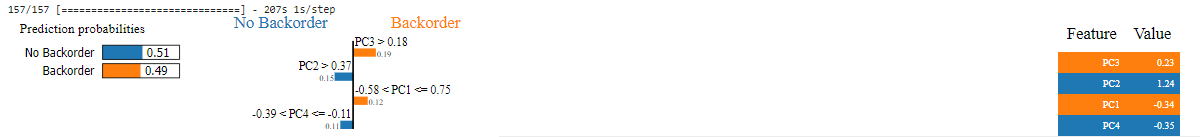}
    \label{lime:subfig1}
  }
  \hfill
  \subfigure[LimeTabularExplainer for a single instane]{
    \includegraphics[width=0.65\textwidth,height=0.4\textheight]{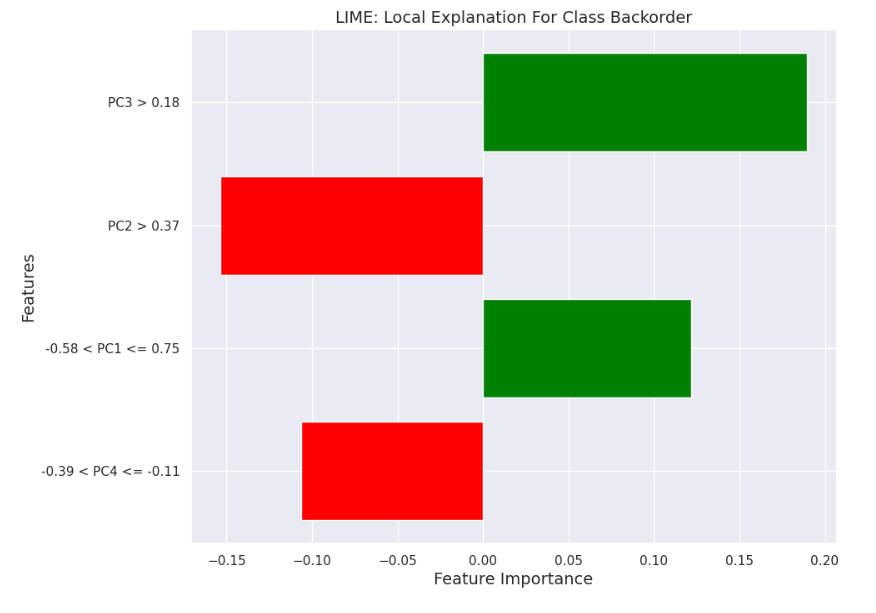}
    \label{lime:subfig2}
  }  
  \caption{The figure comprises two subfigures illustrating the LIME explanations for different instances in the classification task. (a) displays a bar plot depicting the feature importance for a specific instance, while (b) exhibits the LIME-generated explanation plot for another instance. Both (a) and (b) highlight the contributions of the features 'PC1', 'PC2', 'PC3', and 'PC4' towards the predictions, providing insights into the classification process of the model.}
  \label{lime}
\end{figure*}

The family of interpretable models is denoted by $H$, while $\pi_x$ represents the closeness of the instances being evaluated to a specific instance $x$. The term $(h)$ indicates the significance or importance assigned to the model $h$, which can involve additional weighting or importance factors. LimeTabularExplainer is a specific implementation of LIME designed for tabular data. It leverages the idea of perturbation by generating perturbed instances around the instance of interest. LimeTabularExplainer constructs a local surrogate model by fitting a weighted linear model to these perturbed instances. The weights assigned to each perturbed instance reflect their similarity to the original instance, and the model's predictions on these instances are used to approximate feature importance. Figure \ref{lime:subfig1} shows the LIME-based feature importance bar plot showcasing the explanations for a specific instance's prediction. The plot visualizes individual features' contributions towards classifying the instance into "No Backorder" or "Backorder" categories. Figure \ref{lime:subfig2} depicts the LIME-generated explanation plot for another instance, depicting the feature importance and their contributions to the prediction. The features 'PC1', 'PC2', 'PC3', and 'PC4' are considered, and the predicted probabilities are obtained using the model.

\subsubsection*{SHAP}
SHAP is another popular XAI technique that provides global interpretability by attributing the model's predictions to individual features across the entire dataset. Therefore, the contribution of each feature in the prediction was computed and visualized using the SHAP Python library. By utilizing SHAP values from cooperative game theory, SHAP quantifies each feature's influence on a prediction. They provide a quantitative assessment of how much each feature contributes to the overall prediction, denoted as $\phi_j(x)$, which is defined as follows:

\begin{equation}
\phi_j(x) = \frac{1}{m!} \sum_{s \subseteq \{x_1, x_2, ..., x_m\}\setminus\{x_j\}} \frac{|s|!(m - |s| - 1)!}{m!} \times \left( \text{val}(s \cup \{x_j\}) - \text{val}(s) \right)
\label{eq:shap1}
\end{equation}

In Equation \ref{eq:shap1}, $\phi_j(x)$ represents the Shapley value for the feature $x_j$, where $x_j$ denotes a specific feature value. The feature subset of the model is denoted as $s$. The parameter $m$ represents the total number of features in the model. The term $val(s)$ represents the projection of feature values in the set $s$. This equation calculates the Shapley value by summing over all possible subsets $s$, considering their cardinality and the difference between the valuation of the subset including $x_j$ and the valuation of the subset excluding $x_j$. The division by $m!$ and the factorials account for the different permutations and combinations of the subsets.

\begin{figure*}[!ht]
  \centering

  \subfigure[SHAP value visualization for a single prediction]{
    \includegraphics[width=1\textwidth,height=0.05\textheight]{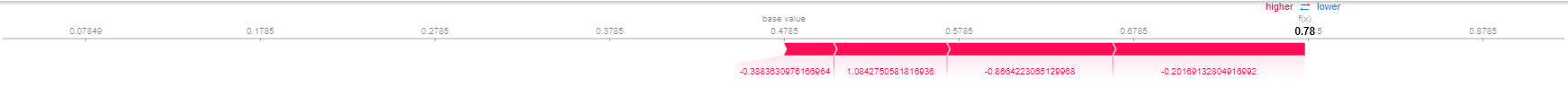}
    \label{shap:subfig1}
  }
  \hfill
  \subfigure[Decision plot of misclassified instances using SHAP values]{
    \includegraphics[width=0.65\textwidth,height=0.28\textheight]{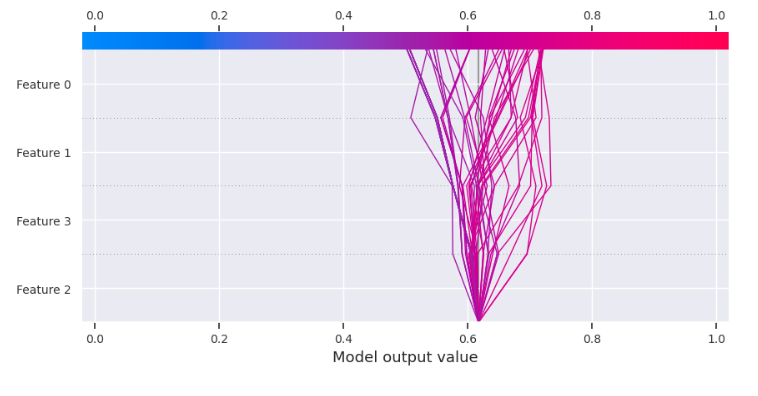}
    \label{shap:subfig2}
  }

  \caption{(a) SHAP force plot on the selected instance using the KernelExplainer and (b) SHAP decision plot showing the features' contributions to the misclassifications in the QAmplifyNet model.}
  \label{shap}
\end{figure*}

We applied SHAP to our hybrid model to understand the importance and influence of different features in determining the model's output. SHAP values explain how the model's expected or base output, denoted by $E[f(x)]$, transitions to the actual output, denoted as $f$ when the specific features $(x)$ are not known. These values quantify each feature's contribution to the prediction and indicate the pattern of connection between the features and the target variable $y$. When the SHAP value of a feature is close to -1 or +1, it substantially impacts the prediction of that data point. On the other hand, a SHAP value close to  0 for a feature indicates less importance in making the prediction. Figure \ref{shap:subfig1} displays the impact of each PC on the prediction for the specific test instance. 

SHAP decision plots provide insights into how the model reaches its decisions, especially when numerous significant features are at play. Each line corresponds to a prediction in these plots and interacts with the x-axis at the predicted value. The color of the line reflects the prediction's value, creating a visual spectrum. Moving from the bottom to the top, SHAP values for each feature are added to the model's baseline, effectively revealing the individual contributions of these features to the overall prediction. These plots are invaluable for understanding the inner workings of complex models by highlighting feature importance and showcasing the impact of each feature on the final prediction. Figure \ref{shap:subfig2} shows how each feature contributes to shifting the model's output from the expected value.

\subsection*{Statistical Test}

We conducted a rigorous statistical analysis with a significance level ($\alpha$) set at 5\% to assess the performance of the QAmplifyNet model against 17 ML models employed in our study. Our research investigated the following two hypotheses:

\begin{itemize}
    \item \textit{Null Hypothesis (H0):} There is no significant difference in accuracy and ROC-AUC between the QAmplifyNet and the tested models.
    \item \textit{Alternative Hypothesis (Ha):} There is a significant difference in accuracy and ROC-AUC between the QAmplifyNet and the tested models.
\end{itemize}

We conducted a ten-fold cross-validated paired t-test, a well-established method for comparing classification model performance. The dataset was randomly split into ten subsets, with shuffling for statistical robustness. Then, for each cross-validation fold, we train the models on the training subset and evaluate their accuracy on the corresponding test subset. Table \ref{tab:stat} displays the results, highlighting significant differences in accuracy and ROC-AUC scores between the QAmplifyNet model and its counterparts. These results reject H0 in all 17 cases, indicating that the observed differences in accuracy and ROC-AUC scores are statistically significant. When the p-value is less than 5\%, it signifies that these differences are unlikely to have occurred by chance, providing robust evidence of the QAmplifyNet model's superior performance. The consistently low p-values underscore our confidence in the model's capabilities.

\begin{table}
\centering
\resizebox{\linewidth}{!}{%
\begin{tabular}{|l |l |l |l |l|} \hline      
\textbf{Model} & \textbf{t-statistic (Accuracy)} & \textbf{p-value (Accuracy)} & \textbf{t-statistic (ROC-AUC)} & \textbf{p-value (ROC-AUC)} \\ \hline 
Catboost & 71.81148356 & 9.95E-14 & 17.63726588 & 2.74E-08 \\ \hline 
DT & 63.94133809 & 2.82E-13 & 17.72554207 & 2.63E-08 \\ \hline 
KNN & 57.87545827 & 6.91E-13 & 73.09019106 & 8.49E-14 \\ \hline 
LGBM & 63.01465217 & 3.22E-13 & 14.14082681 & 1.88E-07 \\ \hline 
RF & 68.07699279 & 1.61E-13 & 14.08814707 & 1.94E-07 \\ \hline 
SVM & 65.40333765 & 2.31E-13 & 34.80494872 & 6.59E-11 \\ \hline 
XGBoost & 71.80695901 & 9.96E-14 & 24.47760546 & 1.52E-09 \\ \hline 
3 Dense Layered NN & 51.25034807 & 2.06E-12 & 11.31959904 & 1.26E-06 \\ \hline 
QSVM+LGBM+LR & 102.2786348 & 4.14E-15 & 26.03690965 & 8.77E-10 \\ \hline 
VQC+QSVM & 116.0140734 & 1.33E-15 & 34.87689033 & 6.47E-11 \\ \hline 
VQC+LGBM & 78.4710231 & 4.49E-14 & 29.79826722 & 2.64E-10 \\ \hline 
MERA 1-Layered & 68.41589025 & 1.54E-13 & 40.62670723 & 1.65E-11 \\ \hline 
MERA 2-Layered & 66.83965127 & 1.90E-13 & 48.78815222 & 3.20E-12 \\ \hline 
MERA 4-Layered & 30.86211809 & 1.93E-10 & 53.36665826 & 1.43E-12 \\ \hline 
RY-CNOT 6-Layered & 23.77391947 & 1.97E-09 & 38.83465142 & 2.47E-11 \\ \hline 
Classical NN+Encoder+QNN & 30.96578613 & 1.87E-10 & 15.72792266 & 7.47E-08 \\ \hline 
DDQN & 62.70623552 & 3.37E-13 & 61.82682204 & 3.82E-13 \\ \hline

\end{tabular}
}
\caption{Ten-fold Cross-Validated Paired t-Test for Model Performance Comparison }
\label{tab:stat}
\end{table}

\section*{Discussions}
\label{sec:discussion}
SCM is a complex and critical process that relies heavily on accurate prediction of backorders to optimize inventory control, reduce costs, and ensure customer experience. This research introduced a groundbreaking hybrid Q-CNN model called QAmplifyNet for SCBP, which integrates quantum-inspired techniques into the conventional ML framework. The discussion aims to comprehensively analyze the proposed model's benefits, limitations, practical implications, and potential applications. The integration of quantum-inspired techniques in the proposed model offers several advantages over classical and hybrid models. Firstly, the utilization of quantum-inspired algorithms enables the model to grasp intricate data patterns and interdependencies, which is crucial for accurate SCBP. The parallelism inherent in QC allows for more efficient solution space exploration, leading to improved prediction accuracy. QAmplifyNet benefits from the flexibility and interpretability of the Keras NN framework. Combining quantum-inspired optimization algorithms with Keras's well-established architecture enhances the model's overall performance and interpretability. Our proposed model demonstrates robustness in handling short, imbalanced datasets commonly encountered in SCM. By employing a combination of preprocessing techniques, undersampling, and principal component analysis, the model effectively addresses the challenges posed by limited data availability and class imbalance.

While QAmplifyNet offers numerous advantages, it is important to acknowledge its limitations. One potential limitation is the computational complexity associated with quantum-inspired techniques. QC is still in its nascent stages, and the current hardware limitations, such as noise and limited qubit connectivity, can hinder the scalability and practical implementation of quantum algorithms. Therefore, the proposed model may face challenges when scaling up to larger datasets or real-time applications. Additionally, the training and optimization of quantum-inspired models require specialized knowledge and expertise. The integration of quantum and classical components in the proposed model adds complexity, requiring researchers and practitioners to have a strong understanding of both QC principles and traditional ML techniques.

Accurate SCBP has significant practical implications for various aspects of SCM. By leveraging the proposed model, organizations can optimize inventory control, reduce backorders, and enhance customer satisfaction. The ability to predict backorders enables proactive management of inventory levels, minimizing stockouts, and ensuring the availability of products to meet customer demands. This, in turn, leads to improved customer loyalty and increased revenue opportunities. The accurate prediction of backorders allows for more efficient resource allocation. Organizations can optimize their production schedules, procurement processes, and transportation logistics based on predicted demand, leading to cost savings and improved operational efficiency. Additionally, accurate SCBP facilitates better supplier communication and coordination, ensuring timely replenishment and minimizing delays.

Our proposed model can be seamlessly integrated into real-world SCM systems. Organizations can enhance their decision-making processes and automate SCBP by incorporating the QAmplifyNet into existing inventory management software. This integration enables real-time monitoring of inventory levels, proactive order fulfillment, and efficient allocation of resources. The model can also be used to identify potential bottlenecks or vulnerabilities in the SC, allowing organizations to implement preventive measures and improve overall SC resilience. QAmplifyNet has the potential for broader applications beyond SCM. The hybrid nature of this model makes it adaptable to other supervised binary classification tasks, such as credit card default prediction or fraud detection, where imbalanced datasets and limited feature sets are common challenges.

The comparative analysis sheds light on the strengths and weaknesses of each model, which has direct implications for SCBP. QAmplifyNet emerges as the top-performing model, consistently demonstrating strong performance across multiple evaluation metrics, including accuracy, F1-score, specificity, Gmean, IBA, and AUC-ROC. Its ability to achieve high accuracy and F1-scores indicates its effectiveness in correctly predicting positive and negative instances, which is crucial for efficient SCM. The superior performance of QAmplifyNet in various metrics implies that it can effectively minimize false positives and false negatives, addressing the challenge of imbalanced data in SCBP. This is particularly noteworthy given the significant impact of FPs and FNs on inventory management and customer satisfaction. Businesses may improve customer satisfaction, reduce the likelihood of disruptions, and maximize inventory efficiency by quickly and correctly detecting instances at risk of backorders.

However, it is essential to consider the practical implications beyond model performance metrics. Factors such as generalizability, interpretability, and computational efficiency are critical for real-world implementation. QAmplifyNet exhibits strong generalization capabilities, as evidenced by its robust performance on the validation dataset. Its incorporation of amplification techniques ensures scalability and computational efficiency, enabling timely predictions for large-scale SC operations. Interpretability is also a crucial factor in SC decision-making. While QAmplifyNet performs exceptionally well in terms of accuracy and other metrics, its black-box nature may limit the understanding of how and why specific predictions are made. To address this, we presented the interpretability of QAmplifyNet using SHAP and LIME.

\section*{Conclusions and Future Works}
\label{sec:conclusion}
In this research, our primary contribution lies in the development of QAmplifyNet, a novel hybrid Q-CNN model designed explicitly for backorder prediction in the SC domain. By harnessing the power of quantum-inspired techniques within the well-established Keras framework, we aimed to significantly enhance the accuracy of backorder prediction. Furthermore, we proposed a comprehensive methodological framework encompassing various stages, including data source identification, data collection, data splitting, data preprocessing, and implementing the QAmplifyNet model. To ensure the optimal performance of our model, we thoroughly explored seven different preprocessing alternatives and meticulously evaluated their effectiveness by assessing the performance of LR on each preprocessed dataset. This rigorous evaluation process allowed us to select the most suitable preprocessing technique for our specific application. Through extensive experiments and evaluations on a short SCBP dataset, we compared the performance of QAmplifyNet with eight traditional CML models, three classically stacked quantum ensemble models, five QNN models, and one deep RL model. Our findings clearly demonstrate the exceptional backorder prediction accuracy achieved by QAmplifyNet, surpassing all other models in terms of accuracy with an impressive 90\% accuracy rate. Notably, QAmplifyNet also achieved the highest F1-score of 94\% for predicting "Not Backorder" and 75\% for predicting "Backorder," outperforming all other models. Additionally, QAmplifyNet exhibited the highest AUC-ROC score of 79.85\%, further validating its superior predictive capabilities. By seamlessly integrating quantum-inspired techniques into our model, we successfully captured complex patterns and dependencies within the data, leading to significant improvements in prediction accuracy.

The significance of the proposed model lies in its ability to optimize inventory control, reduce backorders, and enhance overall SCM. Accurate SCBP enables proactive decision-making, efficient resource allocation, and improved customer satisfaction. By integrating the QAmplifyNet into real-world SC systems, organizations can achieve cost savings, increased revenue opportunities, and improved operational efficiency. By implementing XAI techniques, specifically SHAP and LIME, we could successfully enhance the interpretability of the proposed model. Understanding the model's decision-making process was greatly aided by these XAI techniques, shedding light on the significance and contribution of different features in predicting backorders. By leveraging SHAP and LIME, we were able to gain a deeper understanding of how the model arrived at its predictions and identify the key factors influencing those predictions.

Our pioneering research in quantum-backed backorder prediction presents intriguing future prospects while acknowledging certain technological limitations. We exclusively employed AE as part of our quantum feature mapping. However, the potential of Angle Encoding and Basis Encoding remains untapped, awaiting exploration in subsequent research. Due to constraints in current quantum hardware, our QAmplifyNet model was crafted with only 2 qubits. Yet, as QC technology progresses, the utilization of circuits with a larger qubit capacity, such as 4 or more, holds great promise for tackling more complex computational tasks. Our study featured VQC with a single layer, but the potential for increased depth to capture more nuanced data dependencies beckons; furthermore, by diversifying the types of quantum layers used beyond the SEL, the model's performance can be fine-tuned with the incorporation of Basic Entangler Layers, Continuous-Variable Neural Net Layers, and Random Layers. Additionally, our research primarily integrated VQC with fully connected dense layers. However, there is considerable potential in coupling quantum models with a broader array of neural network architectures, including Long Short-Term Memory, Gated Recurrent Unit, and Convolutional NN. These endeavors are intricately linked to the ongoing evolution of QC capabilities. They will play a vital role in the continued development of QML for diverse and complex applications.

Deploying QAmplifyNet in real-world scenarios presents several potential pitfalls and challenges that must be considered. It's important to acknowledge the current limitations in quantum hardware. Quantum computers are still in their nascent stages, and access to large-scale, fault-tolerant quantum devices is limited. This can restrict the deployment of quantum models in real-world applications, especially when handling extensive datasets or requiring real-time processing. However, with the ongoing advancements in quantum technology, these hardware limitations may gradually diminish. Deeper circuits may demand more computational resources and time, potentially limiting their practicality in real-time applications. Striking a balance between circuit depth and model performance is essential. The adoption of quantum technology can be associated with substantial costs, including hardware, software, and expertise. Assessing the cost-effectiveness of deploying quantum models in real-world scenarios is crucial, particularly for organizations with budget constraints. As QC matures and becomes more accessible, many of these challenges may be addressed, opening up new opportunities for leveraging QML in diverse and complex scenarios.

There are several promising avenues for future work in this field. Firstly, further improvements can be made to the proposed model by exploring additional quantum-inspired techniques and algorithms. As the field of QC continues to advance, more efficient quantum hardware and algorithms are expected to become available, which could enhance the performance and scalability of the model. Expanding the dataset used for training and evaluation could further improve the model's accuracy and generalizability. The model's ability to capture a wider variety of patterns and trends might benefit from the incorporation of a more extensive and diversified dataset. Furthermore, the potential for applying the proposed model in other domains of SCM, such as demand forecasting or inventory optimization, warrants exploration. The versatility of QML models opens up opportunities for their application in various aspects of SC operations.

In addition to introducing a novel strategy for SCBP by making use of the hybrid Q-CNN, this study is notable for being the first use of QML in the field of SCM. The results stress the necessity of quantum-inspired methods to enhance prediction accuracy and optimize SCM. Future research has the potential to change the field of SCM and stimulate breakthroughs in QML models by continuing to improve the model, expanding the dataset, and exploring other quantum-inspired approaches.

\section*{Data availability}

This research used the scikit-learn package for CML trials \cite{scikit}:
\href{https://scikit-learn.org/stable/}{https://scikit-learn.org/stable/}. This is a link to the readily accessible SCBP dataset: \href{https://www.kaggle.com/datasets/gowthammiryala/back-order-prediction-dataset}{\textit{"Can you predict product backorder?"}}.

\section*{Acknowledgements}
We gratefully acknowledge the support of the Competitive Research Fund of The University of Aizu, Japan, for funding this research.

\section*{Author contributions statement}

M.A.J.: Conceptualization, Methodology, Data curation, Writing - Original Draft Preparation, Software, Visualization, Investigation.
M.S.H.S.: Methodology, Writing - Original Draft Preparation.
M.S.I.: Supervision, Writing - Reviewing and Editing.
J.S.: Funding acquisition, Writing - Reviewing and Editing.
M.F.M.: Writing - Reviewing and Editing.
Y.O.: Writing - Reviewing and Editing.

\section*{Competing interests}
The author declares no competing interests.

\section*{Additional information}
Correspondence and requests for materials should be addressed to M.A.J.

\end{document}